\documentclass[fleqn,12pt]{wlscirep}
\usepackage[utf8]{inputenc}
\usepackage[T1]{fontenc}

\usepackage{array}
\newcolumntype{L}[1]{>{\raggedright\let\newline\\\arraybackslash\hspace{0pt}}m{#1}}
\newcolumntype{C}[1]{>{\centering\let\newline\\\arraybackslash\hspace{0pt}}m{#1}}
\newcolumntype{R}[1]{>{\raggedleft\let\newline\\\arraybackslash\hspace{0pt}}m{#1}}

\usepackage{caption}
\usepackage{subcaption}
\usepackage{url}

\usepackage{breakurl}
\usepackage{multirow}
\usepackage{longtable}
\usepackage{tabularx}
\usepackage{makecell}

\usepackage{fancyvrb,fvextra}
\usepackage{setspace}

\usepackage{multibib}
\newcites{supp}{References}

\newcommand{\oursystem}{\textsc{Muse}}
\title{Customized large language models can outperform Community Notes in correcting misinformation}

\author[1,2,*]{Xinyi Zhou}
\author[1,3]{Ashish Sharma}
\author[1,*]{Amy X. Zhang}
\author[1,*]{Tim Althoff}
\affil[1]{Paul G. Allen School of Computer Science and Engineering, University of Washington, Seattle, WA 98195, USA}
\affil[2]{Current affiliation: Computer Science Department, Boise State University, Boise, ID 83702, USA}
\affil[3]{Current affiliation: Microsoft Corporation, Redmond, WA 98052, USA}

\affil[*]{To whom correspondence should be addressed. Email: xinyizhou@boisestate.edu, axz@cs.washington.edu, althoff@cs.washington.edu}

\begin{document}

\flushbottom
\maketitle

\section*{Abstract}
Addressing misinformation in real-world settings is challenging: content is often multimodal; factuality judgments are nuanced and context-dependent; new events emerge rapidly across domains; corrections must be timely, trustworthy, and politically impartial; and multidimensional, multistakeholder frameworks remain lacking. Crowdsourced fact-checking systems such as Community Notes have gained broad adoption, but timely, scalable coverage remains difficult.
We introduce \oursystem, which augments large language models (LLMs) with trust-aware retrieval of up-to-date evidence and task-specific multimodal reasoning. Given a piece of content, \oursystem{} identifies whether and which parts may be false or misleading and provides explanations grounded in credible references. We also develop an evaluation framework that assesses expert-rated response quality---including identification accuracy, explanation factuality, and the relevance and credibility of supporting references---as well as user perceptions.
Across social media posts spanning modalities, domains, political leanings, misinformation tactics, and popularity, \oursystem{} consistently produces high-quality responses, including for content not previously fact-checked online, and outperforms even highly rated Community Notes by 29\%. 
It also improves participants' recognition of misinformation by 10\%.
Our work establishes a general methodological and evaluative framework for timely, scalable, and trustworthy correction of misinformation.

\paragraph{Keywords} Misinformation, large language model, retrieval-augmented generation, multimodality, social media

\section*{Significance Statement}
False and misleading content is pervasive and difficult to address at scale, with serious implications for democratic processes, economic stability, public health, and social cohesion. We present a framework for timely, scalable, and trustworthy correction of misinformation. \oursystem{} identifies potentially misleading claims in multimodal content and explains its assessments using evidence-grounded references from credible publishers, enabling auditable verification. Evaluations with fact-checking experts and laypeople across diverse real-world content demonstrate its effectiveness.

\section*{Introduction}
Misinformation can undermine democratic processes, economic stability, public health, and social cohesion\cite{lazer2018science,zhou2020survey,watts2021measuring,van2022misinformation,moore2023exposure,wef2024risk}. These risks are especially acute on social media, where false content can spread rapidly and the surrounding interface and habitual sharing can impede truth discernment\cite{vosoughi2018spread,epstein2023social,ceylan2023sharing}. 
Correcting misinformation in this setting is difficult: posts may be multimodal; claims can be partly accurate yet misleading through omitted context or spurious causal framing; new events emerge quickly across domains; and useful corrections must be timely, explanatory, evidence-grounded, trustworthy, and politically impartial\cite{watts2021measuring,traberg2022misinformation,dufour2024ammeba,augenstein2023factuality}. Context-rich corrections can improve discernment, whereas delayed or selectively applied corrections have less impact and may create an implied-truth effect for unlabeled content\cite{porter2022factual,van2022misinformation,bak2022combining,pennycook2020implied}.

Professional fact-checkers and crowdsourced systems have become prominent responses to online misinformation, with aggregated crowd judgments often matching professional fact-checkers' assessments\cite{allen2021scaling,martel2022crowds}. X's Community Notes is a leading example: explanatory notes are perceived as more trustworthy than context-free warnings and can reduce subsequent engagement with misleading posts, while citations to external, less politically biased sources are associated with higher helpfulness ratings\cite{wojcik2022birdwatch,drolsbach2024community,slaughter2025community,solovev2025references}. Yet these benefits do not guarantee timely or comprehensive coverage. In our February 2023 snapshot, 88\% of suspicious posts received no response and 93\% received no highly rated response within the first hour (Supp. Fig.~\ref{supp:figure:timeliness_notes}). Scalable methods that preserve the strengths of evidence-rich, auditable notes while reducing these coverage and latency constraints are therefore needed.

LLMs offer a complementary path to scale. Prior work has explored their use in detecting and verifying claims, generating rationales for model judgments, synthesizing crowd-written notes, and drafting full-length fact-checks\cite{yue2024evidence,zeng2024justilm,kim2024can,qi2024sniffer,khaliq2024ragar,de2025supernotes,wu2026beyond,sahnan2026can}. 
These advances are important, but existing systems often address only one stage of the fact-checking workflow or operate under restricted conditions, such as isolated claims, text-only inputs, a single domain or deception type, or reliance on existing fact-checks or crowd-written notes.
Evaluation poses a related challenge. Static misinformation benchmarks can be susceptible to training-data contamination and memorization, making it difficult to assess performance on emerging events\cite{xu2025triplefact}. Moreover, lexical-overlap metrics such as ROUGE, and even automated model-based grading, cannot on their own determine whether a correction accurately identifies the misleading content, provides a factual and complete explanation, and cites relevant and credible sources\cite{augenstein2023factuality}. 
End-user effects likewise cannot be inferred from benchmark accuracy. AI-based dialogues can reduce conspiracy beliefs, but imperfect LLM-generated fact-checks may fail to improve, and may even reduce, headline discernment; user perceptions can also vary depending on whether AI authorship is disclosed\cite{costello2024durably,deverna2024fact,lim2024effect}. 
A practical evaluation should therefore combine expert assessment of complete responses and their supporting evidence with measurement of end-user perceptions.

Here, we introduce \oursystem{} (\underline{mu}ltimodal mi\underline{s}information corr\underline{e}ction), a customized LLM framework for timely, scalable, and trustworthy misinformation correction. \oursystem{} augments LLMs with trust-aware retrieval of up-to-date evidence and task-specific multimodal reasoning. Given a social media post, it identifies (in)accurate and potentially misleading elements and explains its assessment using auditable references from credible sources.
We evaluate \oursystem{} on real posts spanning modalities, domains, political viewpoints, misinformation tactics, and user popularity including posts for which we found no prior online fact-check. Fact-checking experts assess 13 dimensions spanning identification accuracy, explanation factuality, the relevance and credibility of references, and overall response quality; a study with 988 participants assesses end-user perceptions. Across this diverse setting, \oursystem{} achieves an approximately 29\% higher expert-rated response-quality score than highly rated Community Notes. It also improves participants' recognition of misinformation by approximately 10\%.
These findings demonstrate the promise of customized LLMs for correcting real-world misinformation and provide a methodological and evaluative framework for designing and assessing such systems.
\section*{Approach}
  
\begin{figure}
    \centering
    \includegraphics[width=\textwidth]{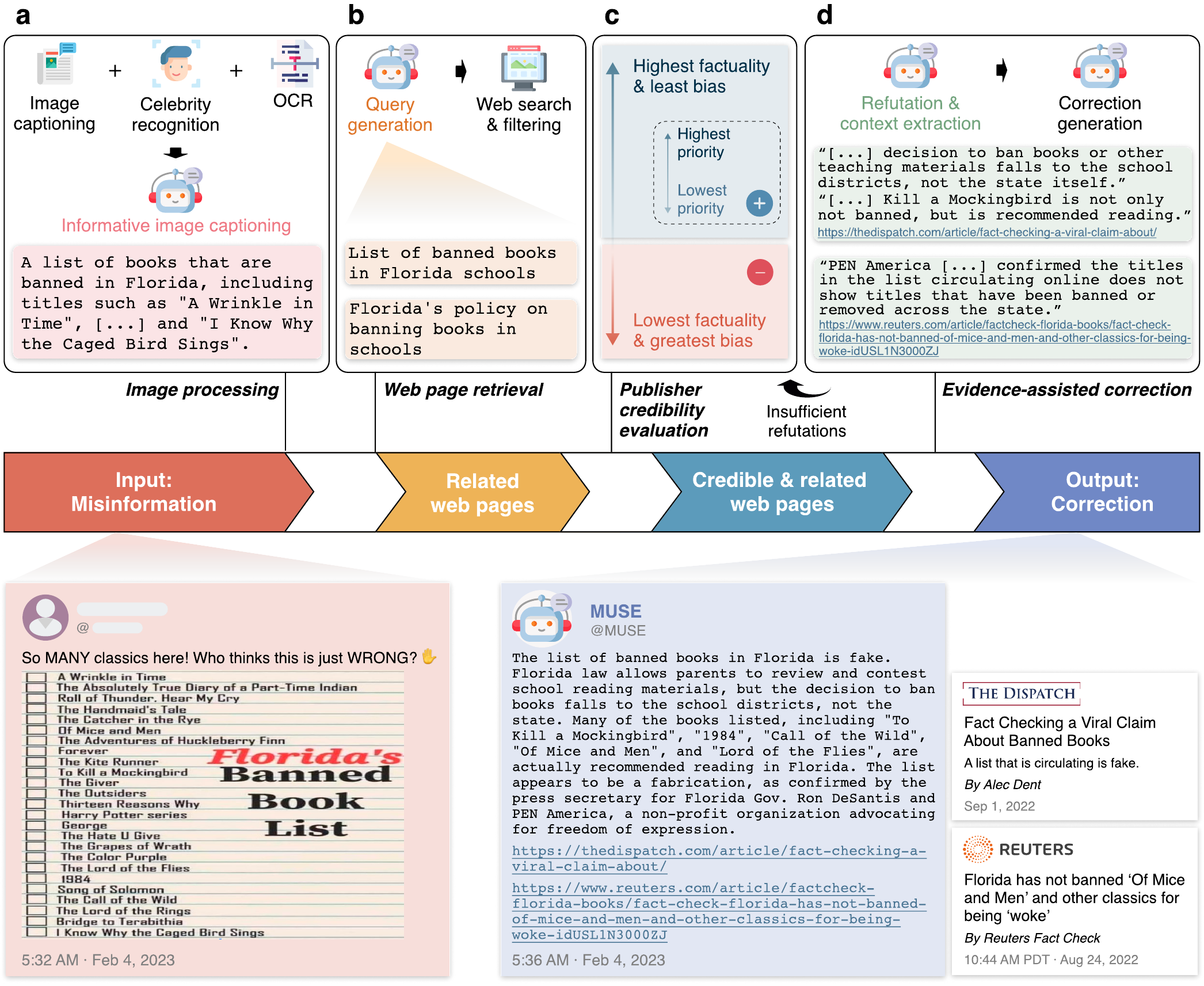}
    \caption{
    Overview of \oursystem, an LLM augmented with trust-aware retrieval of current evidence and task-specific multimodal reasoning to identify and explain (in)accuracies and provide credible supporting references.
    \textbf{a}:~Informative image captioning. \oursystem{} augments image captioning models with celebrity recognition and optical character recognition to produce informative image descriptions.
    \textbf{b}:~Web page retrieval. \oursystem{} uses LLM-generated queries and a web search engine, then filters pages by their multimodal relevance to the content.
    \textbf{c}:~Publisher credibility evaluation.
    \textbf{d}:~Evidence-assisted response generation. \oursystem{} ranks publishers by professionally assessed factuality and bias, then uses an LLM to extract refuting or contextualizing evidence, starting with the most factual and least biased sources. It proceeds down the ranking until finding sufficient refutations (at least two pages) or exhausting all credible pages.}
    \label{fig:1}
\end{figure}

\oursystem{} is an end-to-end system that automatically responds to content containing text, with or without visuals. It identifies accurate and inaccurate claims, explains its judgments, and provides reference links. \oursystem{} comprises three interconnected modules: (1) an LLM for response generation, (2) a hierarchical, credibility-aware web retriever for evidence retrieval, and (3) a multimodal integrator that equips the first two modules with multimodal capabilities. Fig.~\ref{fig:1} illustrates the pipeline. 

Given text-only content, \oursystem{} first generates a set of search queries from the input. Each query is submitted to a web search engine, which returns links to up-to-date web content relevant to the query. After scraping the linked pages, \oursystem{} computes the relevance of each page to the input and removes irrelevant pages. It then evaluates the credibility of the remaining pages using their publishers' factuality and bias ratings and prioritizes pages from publishers with high factuality and low bias. Next, \oursystem{} extracts evidence from the selected pages. This evidence may refute false claims or provide context supporting some or all of the input claims. Finally, \oursystem{} provides the extracted evidence and corresponding source links to an LLM, which generates a response to the input.

Social media content often includes extraneous material that neither requires verification nor contributes to correction, such as unverifiable opinions or emojis, as in Fig.~\ref{fig:1}. Generating targeted queries rather than submitting the post verbatim therefore denoises the input and improves retrieval. Multiple queries also help decompose posts with several claims and reduce incomplete assessments.
Although retaining all retrieved pages could increase the available evidence, it can substantially raise the computational cost of \oursystem{} because of the volume of page content that often needs to be processed. We also find that low-relevance pages increase hallucinations in generated responses (Supp. Fig.~\ref{supp:figure:impact_relevance_computing}).
Moreover, \oursystem{} ranks pages by publisher factuality and bias and extracts evidence from the most factual and least biased sources first. It stops after obtaining sufficient refuting evidence---defined as at least two pages that refute the misinformation---or exhausting all credible pages.

For inputs containing both text and images, \oursystem{} performs informative image captioning to generate a textual description of each image, enabling downstream LLM-based processing. Even when the foundation LLM supports multimodal input, a general-purpose model may not support the identification of public figures (Supp. Fig.~\ref{supp:figure:informative_image_captioning}). Although such restrictions can be motivated by privacy concerns, identifying public figures can be important for misinformation verification and correction. Our end-to-end design seeks to balance these privacy considerations with correction effectiveness.
Specifically, \oursystem{} augments image-captioning models with capabilities that existing models often fail to capture~\cite{stefanini2022show}, including public-figure identification and optical character recognition (OCR). For example, a strong image-captioning model~\cite{li2023blip} may describe the visual misinformation in Fig.~\ref{fig:1} only as ``a list of banned books in Florida,'' omitting the book titles needed for verification and correction. \oursystem{} also performs multimodal web searches and computes multimodal relevance scores to remove irrelevant pages.
\section*{Experiments}

\paragraph{Data}
\oursystem{} was evaluated using X Community Notes, a crowdsourced fact-checking system shown to mitigate misinformation and now expanding to other platforms~\cite{wojcik2022birdwatch,meta2025cn}. Each free-response fact-check, often written by a lay contributor, receives a helpfulness score aggregated from ratings by users with diverse backgrounds, including different political ideologies. Responses with sufficiently high scores are displayed publicly on the corresponding post~\cite{wojcik2022birdwatch} (Supp. Fig.~\ref{supp:figure:example_helpful_note}).
We included 247 posts published between February 2021 and February 2023, each with at least one high- and one average-helpfulness response as of February 2023. Helpfulness scores ranged from $-0.3$ to $0.6$ and were approximately normally distributed (mean: $0.17$, SD: $0.17$; Supp. Fig.~\ref{supp:figure:dist_helpfulness_scores}). We defined high helpfulness as a score $\geq 0.35$; these responses had a mean of $0.44$, above X's suggested threshold of $0.4$~\cite{wojcik2022birdwatch}. Average-helpfulness responses had scores in $[0.05, 0.25)$ and a mean of $0.17$, matching the overall mean.
The expert annotations described below indicated that 48\% of posts contained both accurate and inaccurate or potentially misleading claims, 46\% were inaccurate or potentially misleading, 3\% were fully accurate and not misleading, and the remaining 3\% were unverifiable (Supp.~Fig~\ref{supp:figure:example_muse_responses}).

\paragraph{Baselines}
We compared responses from Community Notes contributors, \oursystem{}, and its variants. For each post, the evaluation set included one high- and one average-helpfulness Community Notes response. Because these responses were written before our evaluation, \oursystem{} could otherwise benefit from retrieving more recently published web pages. We therefore generated two temporally matched responses per post, limiting retrieval to pages published no later than 30 minutes before the corresponding high- or average-helpfulness response was created (Supp. Fig.~\ref{supp:figure:dist_note_times}).
To evaluate \oursystem's ability to immediately respond to misinformation, we also generated one response per post using only web pages published before the post's post time. 
As ablations, we evaluated \oursystem's foundation LLM alone and, for posts containing images, two additional variants: \oursystem$\backslash$vision, which retained credibility-aware retrieval but disabled vision, and \oursystem$\backslash$retrieval, which retained vision but disabled retrieval.

\paragraph{Implementation Details}
Our primary implementation of \oursystem{} used GPT-4, whose September 2021 cutoff for most training data predates over 89\% of the evaluation posts, mitigating the risk of training-data contamination. We also evaluated \oursystem{} with GPT-5.5, a recent multimodal reasoning model, and the open-weight Llama 3 (Discussion). In the prompting-based pipeline, the LLM generated search queries, extracted evidence, produced evidence-grounded responses, and combined BLIP-2 image captions \cite{li2023blip}, Amazon Rekognition celebrity labels, and Amazon Textract OCR outputs into informative image descriptions using in-context learning (Supp.~Figs.~\ref{supp:figure:prompt_query_generation}--\ref{supp:figure:prompt_image2text}). Google Search served as the web retriever.

To estimate relevance, we encoded the text of each input item and candidate webpage using the pretrained Sentence Transformer \texttt{msmarco-distilbert-base-tas-b}\cite{reimers2019sentence,hofstatter2021efficiently} and their visual content using the pretrained Vision Transformer \texttt{dino-vitb8}\cite{caron2021emerging}, then computed textual and visual similarity scores. We calibrated the relevance thresholds on 100 webpage--input pairs comprising the top 10 search results for each of 10 randomly selected input items excluded from the evaluation set. We manually labeled the pairs and selected thresholds yielding no false-positive inclusions in the calibration set.

We operationalized publisher-level factuality and bias using human-curated ratings from Media Bias/Fact Check, a media-rating organization with broad coverage that has been used in prior research\cite{weld2021political,bozarth2020higher}. We retained publishers rated ``very high,'' ``high,'' or ``mostly factual'' and classified as ``least biased,'' ``left-center,'' ``right-center,'' or ``pro-science'' (Supp.~Table~\ref{supp:table:mbfc}). We then assigned the \(2{,}445\) eligible publishers to three priority tiers: high (``very high'' factuality and ``least biased'' or ``pro-science''; \(n=118\); e.g., CDC, Nature, Pew Research Center), medium (remaining publishers with at least ``high'' factuality; \(n=2{,}123\); e.g., Britannica, Statista), and low (remaining eligible publishers; \(n=204\); e.g., Forbes, USA Today).

\paragraph{Expert Evaluation}

With support from Hacks/Hackers, an international grassroots journalism organization, we recruited fact-checking experts through its email list. Of 15 respondents, we selected the 12 with the most relevant experience and at least fluent English proficiency. Five (42\%) had 1--3 years of domain experience, three (25\%) had 4--6 years, one (8\%) had 7--9 years, and three (25\%) had 9+ years. Nine (75\%) were native English speakers, and three (25\%) were fluent. The UW IRB determined the study exempt (STUDY00017831).

The study had two phases. In Phase I, we remotely explained the annotation protocol (Supp.~Fig.~\ref{supp:figure:annotation_instructions_1}--\ref{supp:figure:annotation_instructions_7}), demonstrated the annotation interface (Supp.~Fig.~\ref{supp:figure:annotation_task_page}), and asked participants to complete three practice tasks with explanations. We reviewed their responses and provided feedback to resolve misunderstandings. Two participants withdrew, and all Phase-I data were excluded from analysis. We divided the remaining ten participants into five pairs and randomly assigned each 26 or 27 Phase-II tasks. Seven tasks (approximately 30\%) overlapped within each pair to measure inter-annotator agreement. All ten completed Phase II and received a \$450 Amazon gift card.

Recruited experts evaluated responses to the same post across four dimensions. For \textit{identifying and explaining factual (in)accuracies}, they assessed explicitness, identification correctness and comprehensiveness, and explanation accuracy and informativeness. For \textit{generated-text quality}, they rated relevance, factuality, fluency\cite{yuan2021bartscore,fabbri2021summeval}, coherence\cite{yuan2021bartscore}, and toxicity. For \textit{reference quality}, they assessed reachability, relevance to the generated text, and credibility. Finally, considering all 13 criteria, they rated \textit{overall correction quality} on a 10-point scale. Evaluators were blinded to the approach used to generate each response.

We measured inter-rater agreement using weighted Cohen's $\kappa$, except for toxicity, fluency, and reference relevance, whose skewed labels warranted observed agreement (0.96, 0.86, and 0.81)\cite{xu2014interrater}. Under standard interpretations\cite{mchugh2012interrater}, agreement was substantial for reference reachability (0.79), moderate for overall quality, informativeness, comprehensiveness, and text relevance (0.41–0.51), and fair for all other $\kappa$-scored criteria (0.28–0.40), consistent with known expert disagreement in misinformation assessment\cite{allen2021scaling,bhuiyan2020investigating}.

\paragraph{Measuring End User Perceptions of Corrections}

Previous research suggests that bias against AI-generated content can reduce its effectiveness and perceived trustworthiness\cite{lim2024effect}. To assess \oursystem's potential real-world impact, we conducted a Prolific study with 988 U.S. participants, following Costello et al.~\cite{costello2024durably} (Supp. Fig.~\ref{supp:figure:user-study-interface-1}--\ref{supp:figure:user-study-interface-2}). The sample was representative of the U.S. population by sex, age, and political affiliation. Each participant viewed a social media post from the expert-evaluation dataset and a corresponding response, then rated their belief in and intention to share the post before and after the response, as well as the response's trustworthiness (all on 1--7 scales). They were randomly assigned to either no source disclosure or explicit disclosure that the correction was human-written or AI-generated.

\section*{Results}

\begin{figure}
    \centering
    \includegraphics[width=\textwidth]{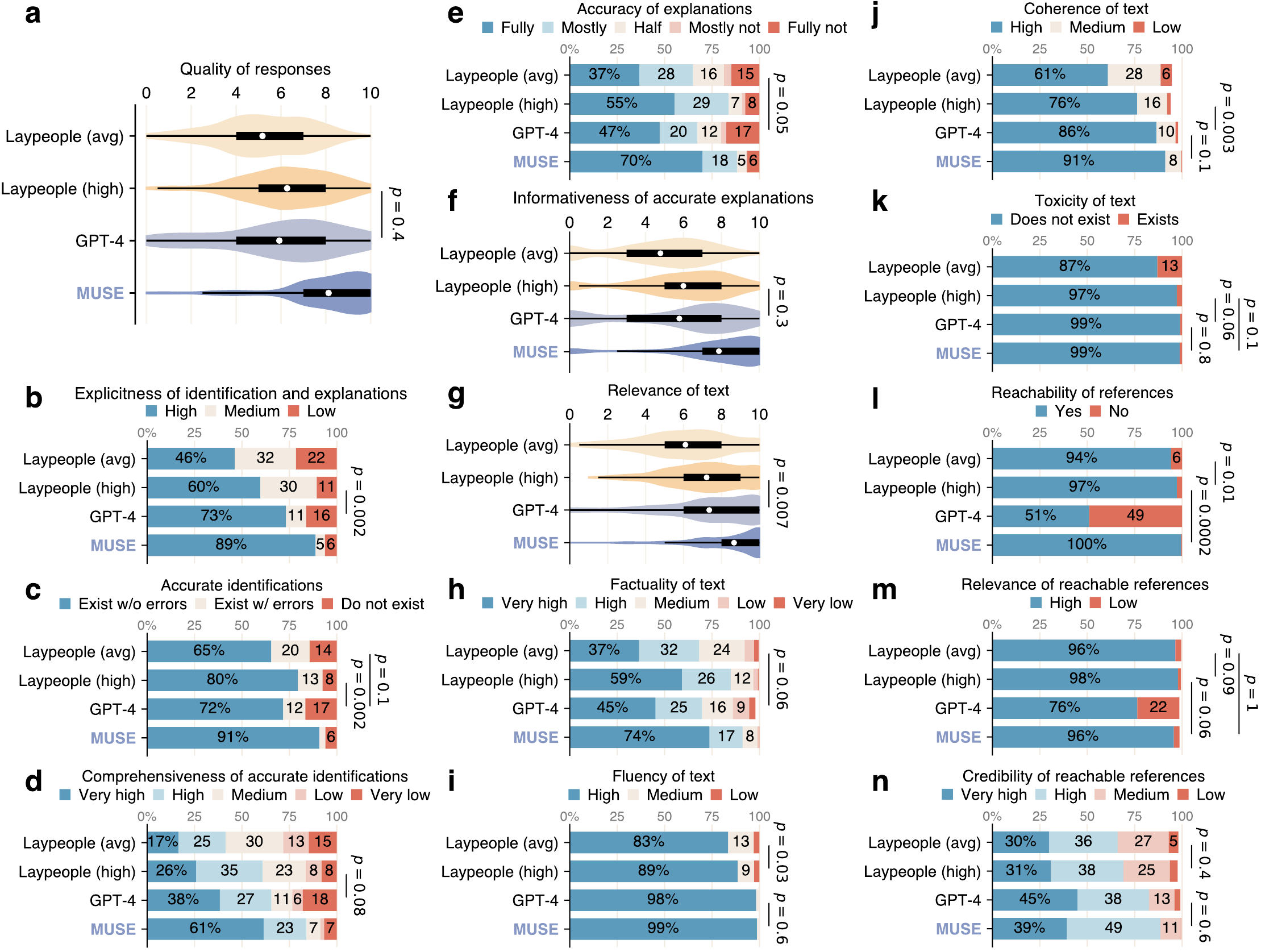}
    \caption{Results of expert evaluation (all pairwise approach comparisons in \textbf{a}--\textbf{n}: Mann--Whitney U test, $p<2\times10^{-5}$; 84 experiments).
    \textbf{a}:~Overall response quality. \oursystem-generated responses ($\text{mean}\pm\text{SD}$: $8.1\pm2.0$; $n=232$) scored 29\%, 37\%, and 56\% higher than laypeople's high-helpfulness responses ($6.3\pm2.0$; $n=232$), GPT-4 responses ($5.9\pm2.7$; $n=232$), and laypeople's average-helpfulness responses ($5.2\pm2.1$; $n=230$), respectively.
    \textbf{b}--\textbf{f}:~Identification and explanation of inaccuracies. Compared with all baselines, \oursystem{} identifies and explains inaccuracies more explicitly (\textbf{b}) and comprehensively, makes fewer accuracy misclassifications (\textbf{c}--\textbf{d}), and provides more accurate and informative explanations (\textbf{e}--\textbf{f}).
    \textbf{g}--\textbf{k}:~Text quality. \oursystem-generated text is more relevant and factual than all baselines (\textbf{g}--\textbf{h}), more fluent and coherent than laypeople's high-helpfulness responses, and less toxic than their average-helpfulness responses (\textbf{i}--\textbf{k}).
    \textbf{l}--\textbf{n}:~Reference quality. Unlike GPT-4, \oursystem{} rarely hallucinates references and provides more reachable, relevant links (\textbf{l}--\textbf{m}); its references are also more credible than those in laypeople's responses (\textbf{n}).}
    \label{fig:2}
\end{figure}

\paragraph{\oursystem-generated responses exceed even high-helpfulness Community Notes in overall quality.}
\oursystem{} achieved a mean quality score of 8.1/10, exceeding laypeople's high-helpfulness responses by 29\% (6.3; $p=3\times10^{-48}$; $N=464$), GPT-4 by 37\% (5.9; $p=4\times10^{-42}$; $N=464$), and laypeople's average-helpfulness responses by 56\% (5.2; $p=5\times10^{-81}$; $N=462$; Mann--Whitney U tests unless noted; Fig.~\ref{fig:2}a). 
Here, lay responses were written an average of 14 hours after posting, whereas \oursystem{} used only web pages published before the post.

\paragraph{\oursystem{} better identifies and explains (in)accuracies.}
Overall, 89\% of \oursystem{} responses explicitly identified and explained (in)accuracies, 16\% more than GPT-4, 29\% more than high-helpfulness notes, and 43\% more than average-helpfulness notes (Fig.~\ref{fig:2}b). Moreover, 91\% contained at least one correct identification and no mistakes, compared with 80\%, 72\%, and 65\%, respectively (Fig.~\ref{fig:2}c); 61\% identified all (in)accuracies, versus 38\%, 26\%, and 17\% (Fig.~\ref{fig:2}d). Fully accurate explanations occurred in 70\% of \oursystem{} responses, compared with 55\% for high-helpfulness notes, 47\% for GPT-4, and 37\% for average-helpfulness notes (Fig.~\ref{fig:2}e). Its mean informativeness score was 7.9, respectively 32\%, 36\%, and 65\% higher (Fig.~\ref{fig:2}f).

\paragraph{\oursystem{} generates higher-quality text.}
By augmenting GPT-4 with timely web access and visual analysis, \oursystem{} improved relevance ($p=10^{-15}$; $N=460$) and factuality ($p=2\times10^{-20}$; $N=459$) without reducing fluency ($p=0.6$; $N=464$), coherence ($p=0.1$; $N=459$), or safety with respect to toxicity ($p=0.8$; $N=464$). Compared with high-helpfulness lay responses, its text was more relevant ($p=2\times10^{-30}$; $N=464$), factual ($p=4\times10^{-6}$; $N=463$), fluent ($p=2\times10^{-10}$; $N=464$), and coherent ($p=10^{-5}$; $N=451$); it was also less toxic than average-helpfulness responses ($p=4\times10^{-12}$; $N=462$). \oursystem{} attained a mean relevance score of 8.7---18\% above GPT-4, 21\% above high-helpfulness notes, and 43\% above average-helpfulness notes (Fig.~\ref{fig:2}g). Its text was completely factual in 74\% of cases, versus 59\% for high-helpfulness notes and 45\% for GPT-4 (Fig.~\ref{fig:2}h); nearly all responses were grammatically correct and free of bias, impoliteness, or provocation, and 91\% were highly coherent and logical (Fig.~\ref{fig:2}i--k).

\paragraph{\oursystem{} provides higher-quality references.}
Among GPT-4's references, 49\% returned page-not-found errors, and only 76\% of reachable links were relevant (Fig.~\ref{fig:2}l--m). For \oursystem{}, nearly 100\% of links were reachable and 96\% were relevant. Its references were also more credible than those in high-helpfulness lay responses ($p=4\times10^{-11}$; $N=744$; Fig.~\ref{fig:2}n).

\paragraph{\oursystem{}'s superior performance is not driven by potential training-data contamination or reliance on existing fact-checks.}

\begin{figure}
    \centering
    \includegraphics[width=\textwidth]{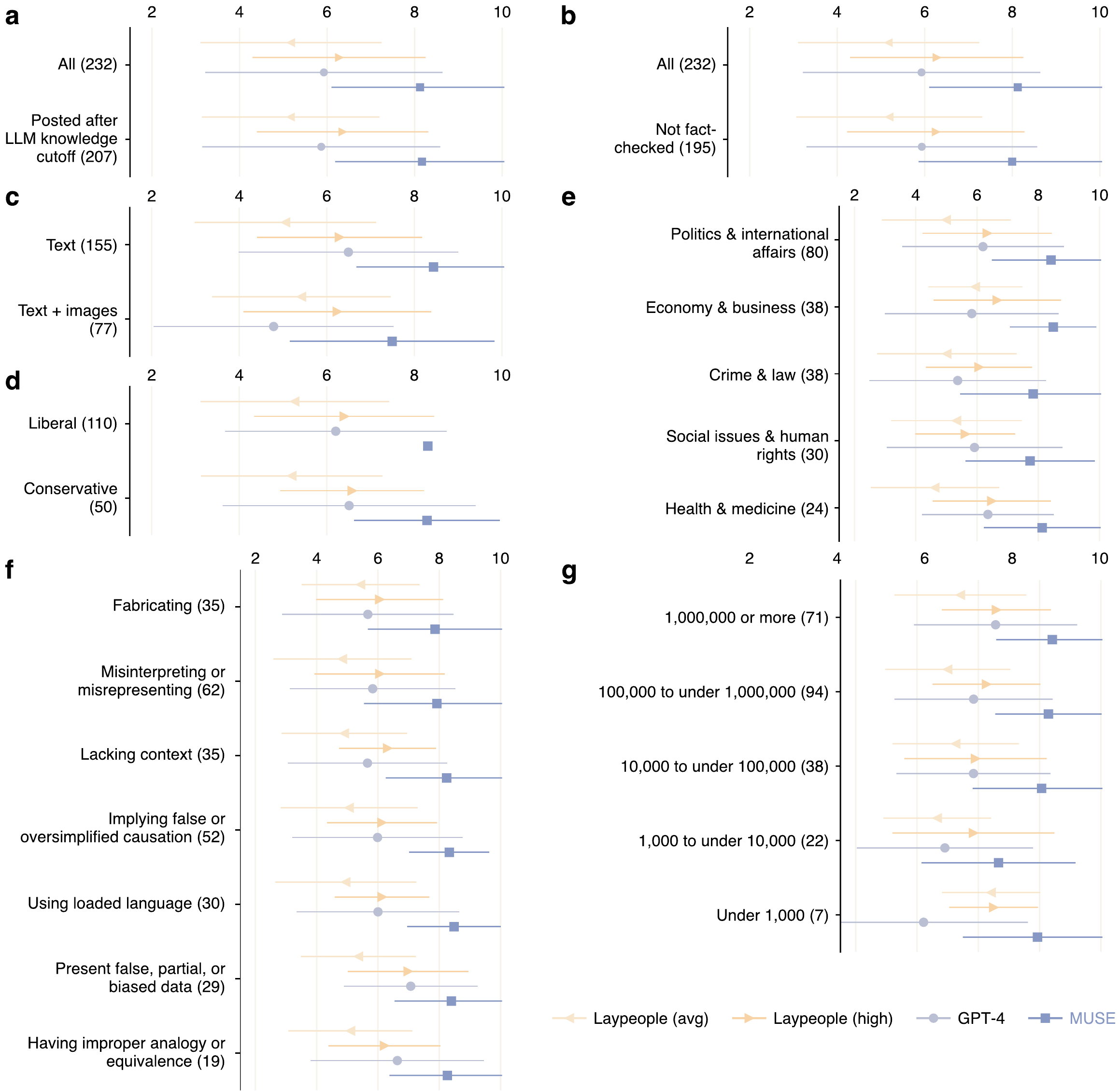}
    \caption{Robustness analyses of \oursystem{}.
    \textbf{a--b}: The superior performance of \oursystem{} does not appear to be driven by potential training-data contamination or reliance on existing fact-checks.
    \textbf{c--g}: Compared with all baselines, \oursystem{} consistently generates higher-quality responses across modalities, political leanings, domains, misinformation tactics, and user popularity.}
    \label{fig:3}
\end{figure}

Using September 2021 as GPT-4's assumed training-data cutoff, we compared performance on all tweets ($n=232$) with performance on tweets posted after the cutoff ($n=207$). Posts and Community Notes from before or during September 2021 might have appeared in GPT-4's training data, potentially inflating performance in ways that would not generalize to future content. Nevertheless, \oursystem{} performed stably on both sets (mean$\pm$SD correction quality: 8.1$\pm$2.0) and consistently outperformed GPT-4 and even high-helpfulness lay responses (Fig.~\ref{fig:3}a).

\oursystem{} cited at least one related fact-checking article for 16\% of all posts and relied exclusively on such articles for only 6\%. After excluding all posts for which \oursystem{} cited a related fact-checking article, its mean response-quality score remained 8.0, close to its mean of 8.1 across all posts. For these posts, the quality of responses generated by \oursystem{} was at least 28\% higher than that of GPT-4 and high-helpfulness responses by laypeople (Fig.~\ref{fig:3}b). Thus, \oursystem{}'s advantage cannot be explained primarily by its ability to summarize retrieved professional fact-checks.

\paragraph{\oursystem{} performs robustly across modalities, political leanings, domains, misinformation tactics, and source popularity.}

For both text-only and multimodal posts, the quality of \oursystem-generated responses exceeded that of GPT-4 and even high-helpfulness lay responses by at least 21\% (Fig.~\ref{fig:3}c). We also annotated the political leaning of each post. GPT-4, which has previously been reported to exhibit a liberal bias~\cite{feng2023pretraining}, scored 5\% lower on liberal-leaning misinformation than on conservative-leaning misinformation (Fig.~\ref{fig:3}d). By contrast, \oursystem{} achieved the same mean quality score of 8.3 for both groups and outperformed GPT-4 and high-helpfulness lay responses by at least 26\%.
We further annotated the topic domain of each post and the tactics that made it, or parts of it, false or potentially misleading. Across domains and misinformation tactics, responses generated by \oursystem{} maintained a mean quality score of approximately 8 and outperformed GPT-4 and even high-helpfulness responses by laypeople by at least 19\% (Fig.~\ref{fig:3}e--f; see example posts using various tactics in Supp.~Fig.~\ref{supp:figure:example_misinfo_tactic}).

Finally, we measured poster popularity by follower count. \oursystem's response quality did not vary significantly with popularity and consistently exceeded that of GPT-4 and even high-helpfulness lay responses (Fig.~\ref{fig:3}g). Community Notes overrepresents relatively popular sources and, like any existing dataset, does not capture the full range of misinformation, including content from obscure accounts. However, it enables the study of high-reach misinformation, which may be especially persuasive and harmful\cite{traberg2022misinformation} yet remains understudied. More broadly, no misinformation dataset is fully representative because selection effects shape both what content is created and what is annotated. Some biases are substantively meaningful: popular accounts reach larger audiences and may warrant faster assessment. Community Notes is therefore an important public dataset reflecting how social media platforms address potential misinformation\cite{meta2025cn}.

\paragraph{Ablation studies demonstrate the value of retrieval and multimodal integration.}

GPT-4 can be viewed as an unaugmented variant of \oursystem{} without the proposed retrieval and vision capabilities. Further ablations show that both components contribute substantially: \oursystem{} outperformed the variant without retrieval by 25\% and the variant without vision by 33\% in overall response quality (Supp.~Fig.~\ref{supp:figure:impact_retrieval_and_vision}).

\paragraph{Correction timing does not affect \oursystem's response quality.}
For each post, we generated three \oursystem{} responses simulating different start times. Performance was similar (mean$\pm$SD response quality: 8.1$\pm$2.0) whether correction began immediately after posting or at the median response times of laypeople producing high-helpfulness notes (13 hours) or average-helpfulness notes (16 hours; Supp.~Figs.~\ref{supp:figure:impact_time_correction} and \ref{supp:figure:dist_note_times}).

\paragraph{\oursystem{} improves misinformation recognition regardless of source disclosure.}

\begin{figure}
    \centering
    \includegraphics[width=\textwidth]{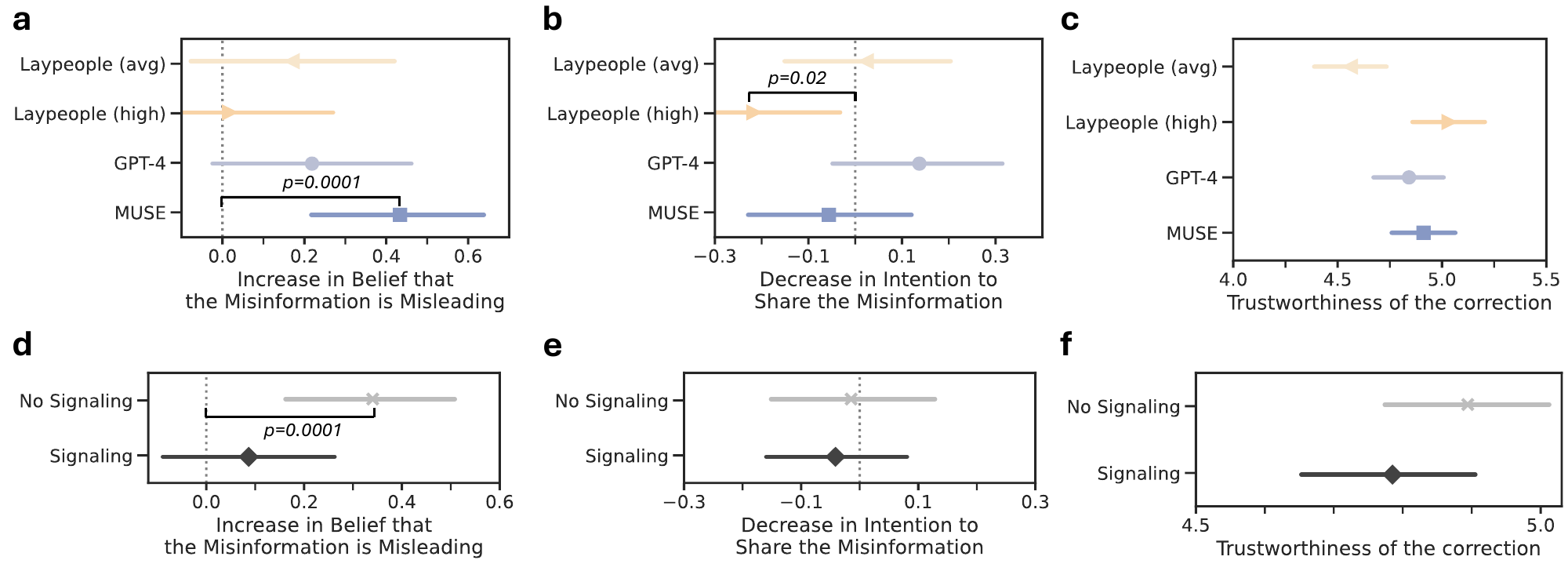}
    \caption{Prolific study with 988 participants to assess the real-world impact of \oursystem{} and how people perceive and react to AI-generated responses (two-sided Student's t-test; 16 experiments). 
    \textbf{a}: Only \oursystem-generated corrections significantly enhanced the correct belief that the misinformation is misleading ($p=0.0001$). 
    \textbf{b}: None of the corrections made by laypeople, GPT-4, and \oursystem{} significantly impacted participants' intention to share the misinformation, and participants' intentions were generally low (around 2/7).
    \textbf{c}: \oursystem, GPT-4, and high-helpfulness responses from laypeople all received significantly higher trustworthiness ratings than average-helpfulness responses from laypeople. (d-f) While disclosing the source of the correction (human or AI) had no significant effect on trustworthiness or intention to share, not disclosing the source significantly enhanced the correct belief that the misinformation is misleading ($p=0.0001$).}
    \label{fig:4}
\end{figure}

\oursystem-generated corrections significantly increased end-user participants' correct belief that the misinformation was misleading by 10\% (from 4.5 to 4.9; Cohen's $d=0.3$; $p=0.0001$; Fig.~\ref{fig:4}a). No significant belief changes were observed for GPT-4 or lay responses of either helpfulness level. 
None of the corrections significantly affected participants' intention to share the misinformation, which was generally low (approximately 2/7; Fig.~\ref{fig:4}b).
\oursystem, GPT-4, and high-helpfulness lay responses were all rated significantly more trustworthy than average-helpfulness lay responses (Fig.~\ref{fig:4}c). \oursystem{} received a mean trustworthiness score of 4.9, 8\% higher than average-helpfulness lay responses (mean: 4.6; $p=0.004$); the corresponding differences were 6\% for GPT-4 ($p=0.03$) and 10\% for high-helpfulness lay responses ($p=0.0002$).

Source disclosure (human or AI) did not significantly affect trustworthiness or sharing intention (Fig.~\ref{fig:4}e--f). However, withholding the source increased participants' correct belief by 8\% (from 4.4 to 4.7; Cohen's $d=0.2$; $p=0.0001$; Fig.~\ref{fig:4}d), primarily for the less effective GPT-4 and average-helpfulness lay responses. By contrast, \oursystem{} and high-helpfulness lay responses were similarly effective regardless of disclosure. Although prior work reports bias against disclosed AI-generated content\cite{lim2024effect}, our results suggest that highly effective corrections remain robust to source disclosure.
\section*{Discussion}

As Community Notes-like approaches are increasingly adopted by social media platforms\cite{wojcik2022birdwatch,meta2025cn}, we propose \oursystem{} as a practical end-to-end framework for helping users and platforms respond to misinformation reliably, promptly, and transparently. Its simple, training-free pipeline facilitates deployment and maintenance. Despite this simplicity, expert evaluation shows that \oursystem{} produces high-quality corrections that accurately identify and explain inaccuracies, generate effective text, and provide credible supporting references, significantly outperforming even highly rated lay responses across misinformation modalities, political leanings, topic domains, tactics, popularity levels, and claims not previously fact-checked online.

In addition to achieving the highest accuracy and factuality, \oursystem{} produces the most readable responses, with greater explicitness, fluency, and coherence. Its low-toxicity text and credible references may reduce correction backfire effects\cite{he2023reinforcement,ecker2022psychological,swire2020searching}, while identifying inaccuracies comprehensively and at scale may mitigate the implied truth effect\cite{pennycook2020implied}. We further show that \oursystem-generated responses improve people's recognition of misinformation, regardless of whether their source is disclosed. 
These corrections can be generated within minutes of suspicious content appearing: \oursystem{} responds to a post in two minutes on average on a basic M1 CPU setup with 16 GB of memory. Our primary design objective was to maximize correction quality. By transparently linking retrieved evidence, \oursystem{} enables users to verify responses while signaling insufficient evidence, providing accuracy nudges, and expressing uncertainty when appropriate (Supp. Fig.~\ref{supp:figure:example_no_evidence}).

Responses generated by \oursystem{} cost approximately 0.2 USD per post, with correction quality as our primary design objective. Given the task's complexity and the substantial quality improvements, this cost is modest relative to alternatives. For example, crowdsourced lay judgments cost about 0.9 USD per article merely to assess whether its headline and lede contain misinformation, without providing an explanation or correction\cite{allen2021scaling}.
We also replaced GPT-4 with the open-weight Llama 3 70B model. Holding all other implementation details fixed, the variants performed comparably in our qualitative analysis (Supp. Fig.~\ref{supp:figure:impact_foundation_llm}), suggesting that \oursystem{} is not tied to a specific foundation model and may support lower-cost deployment.

We further replaced GPT-4 with GPT-5.5, an advanced multimodal reasoning model, and found that such models alone remain insufficient. Vanilla GPT-5.5 still hallucinated references: only 70\% of links were reachable, based on an HTTP check that achieved an F1 score of 0.91 on expert-annotated URLs. Of the reachable links, 81\% were relevant and supported the associated claims, yielding a reference-hallucination rate of 43\%, vs 2\% for \oursystem{}. We assessed relevance by embedding the generated text and crawled reference content using OpenAI's \texttt{text-embedding-3-small} and classifying each pair by cosine similarity. A threshold of 0.1 achieved the highest validation F1 score of 0.86 on expert-annotated URLs.
Although recent commercial models integrate web search, they remain opaque about which sources are retrieved and how evidence is selected, limiting trustworthiness\cite{solovev2025references}.
Llama 3 and GPT-5.5 have knowledge cutoffs of December 2023 and 2025, respectively, so their training data may include evaluation posts, corrections, or Community Notes. Direct comparisons among Llama 3-, GPT-5.5-, and GPT-4-based versions of \oursystem{} are therefore not strictly fair. Our goal is not to propose a state-of-the-art LLM, but to develop a scalable misinformation-response system whose corrections outperform or are comparable to highly rated Community Notes. Stronger future models may make \oursystem{} even more effective.

This work has several limitations. First, \oursystem{} supports English text and image inputs but not video. Second, we evaluated it on X, a single but widely used platform where many users consume news and false information can spread faster than truth\cite{walker2021news,vosoughi2018spread}. We selected X because its transparent Community Notes system has been shown to reduce misinformation diffusion\cite{wojcik2022birdwatch,drolsbach2024community,slaughter2025community}.
Third, as a retrieval-based framework, \oursystem{} depends on sufficiently relevant and credible online evidence. In information voids, such as breaking news, insufficient evidence may prevent reliable verification; the system should therefore communicate uncertainty rather than make definitive judgments. Although Supp. Fig.~\ref{supp:figure:example_no_evidence} illustrates this behavior, we did not systematically evaluate information-void detection or abstention reliability. 
Fourth, our modest Community Notes sample consisted largely of posts containing inaccurate or potentially misleading elements, while often alongside accurate claims, limiting broader generalizability and assessment of truth discernment; end-user outcomes were immediate self-reports\cite{guay2023think}. Larger expert evaluations and field studies are needed.
\bibliography{_references}

\newpage
\section*{Acknowledgements}
We would like to thank Hacks/Hackers for advertising the study and helping with the recruitment. We also thank members of the UW Social Futures Lab and Behavioral Data Science Group for their suggestions and feedback. This work was supported in part by the National Science Foundation's Convergence Accelerator program under Award No. 49100421C0037. T.A. and A.S. were supported in part by the Office of Naval Research (\#N00014-21-1-2154), NSF IIS-1901386, and NSF CAREER IIS-2142794.

\section*{Supplementary Material} 

Supplementary material is available for this paper.

\section*{Author Contributions}
X.Z., A.X.Z., and T.A. conceptualized the work and developed the overarching research goals and aims.
X.Z. led the model development.
A.S. led the assessment of end-user perceptions of corrections, and X.Z. led the remainder of the evaluation.
X.Z. led the manuscript writing, including visualization, and all authors revised the paper.
A.X.Z. and T.A. supervised the study.

\section*{Data Availability}
Data, including code, used in this study are available at \url{https://github.com/Social-Futures-Lab/MUSE}. We comply with X Terms of Service by only releasing the IDs of tweets. The experts' names are anonymized.
\appendix
\newpage

\setcounter{table}{0}
\renewcommand{\thetable}{S\arabic{table}}
\setcounter{figure}{0}
\renewcommand{\thefigure}{S\arabic{figure}}

\section*{Supplementary Materials}
\begin{figure}
    \centering
    \includegraphics[width=0.7\textwidth]{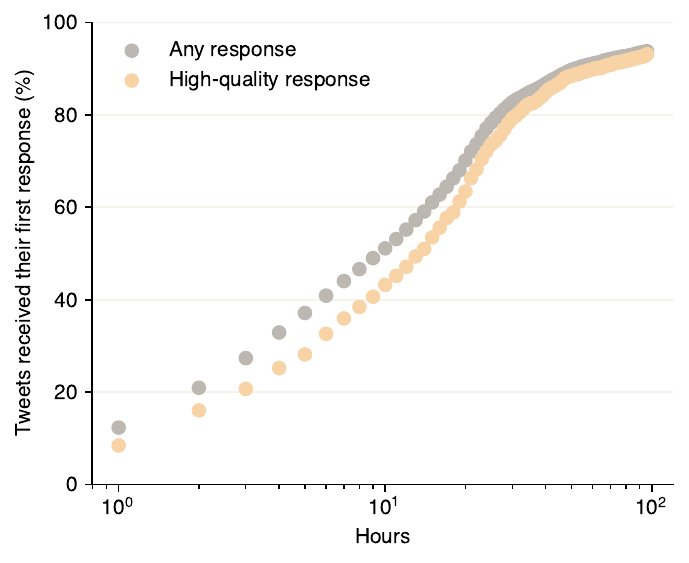}
    \caption{Distribution of potential misinformation in X Community Notes (as of February 2023) that received its first response (gray) or first high-quality response (orange) within a certain amount of time.}
    \label{supp:figure:timeliness_notes}
\end{figure}

\begin{figure}
    \centering
    \includegraphics[width=\textwidth]{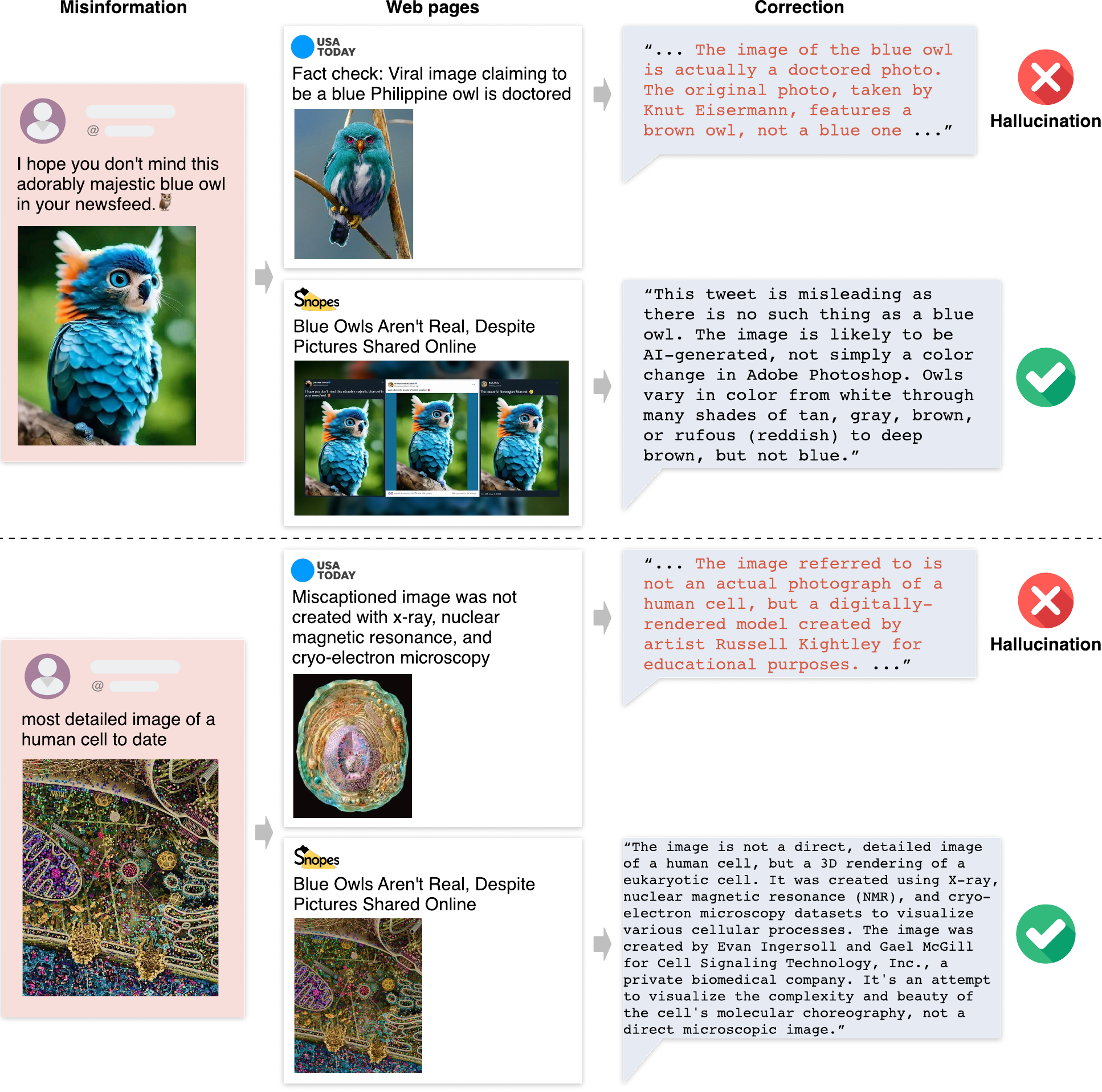}
    \caption{Examples that show how retrieved web pages with relatively low relevance to potential misinformation can promote LLM (in this case, GPT-4) hallucinations when generating responses.}
    \label{supp:figure:impact_relevance_computing}
\end{figure}

\begin{figure}
    \centering
    \includegraphics[width=\textwidth]{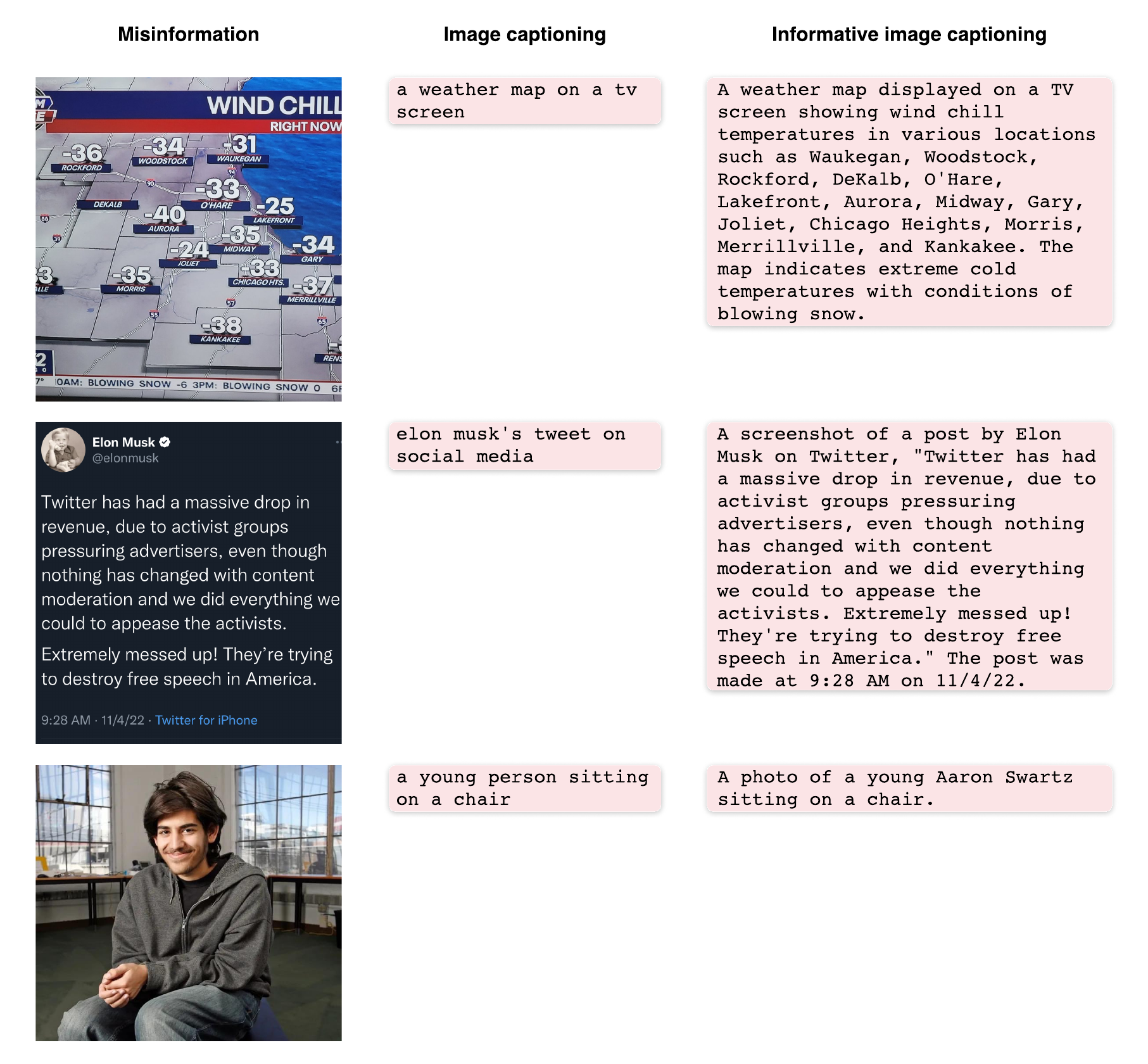}
    \caption{Examples of informative image captions, which augment image captions with names of visually represented celebrities and embedded text (see Experiments for implementation details).}
    \label{supp:figure:informative_image_captioning}
\end{figure}

\begin{figure}
    \centering
    \includegraphics[width=.7\textwidth]{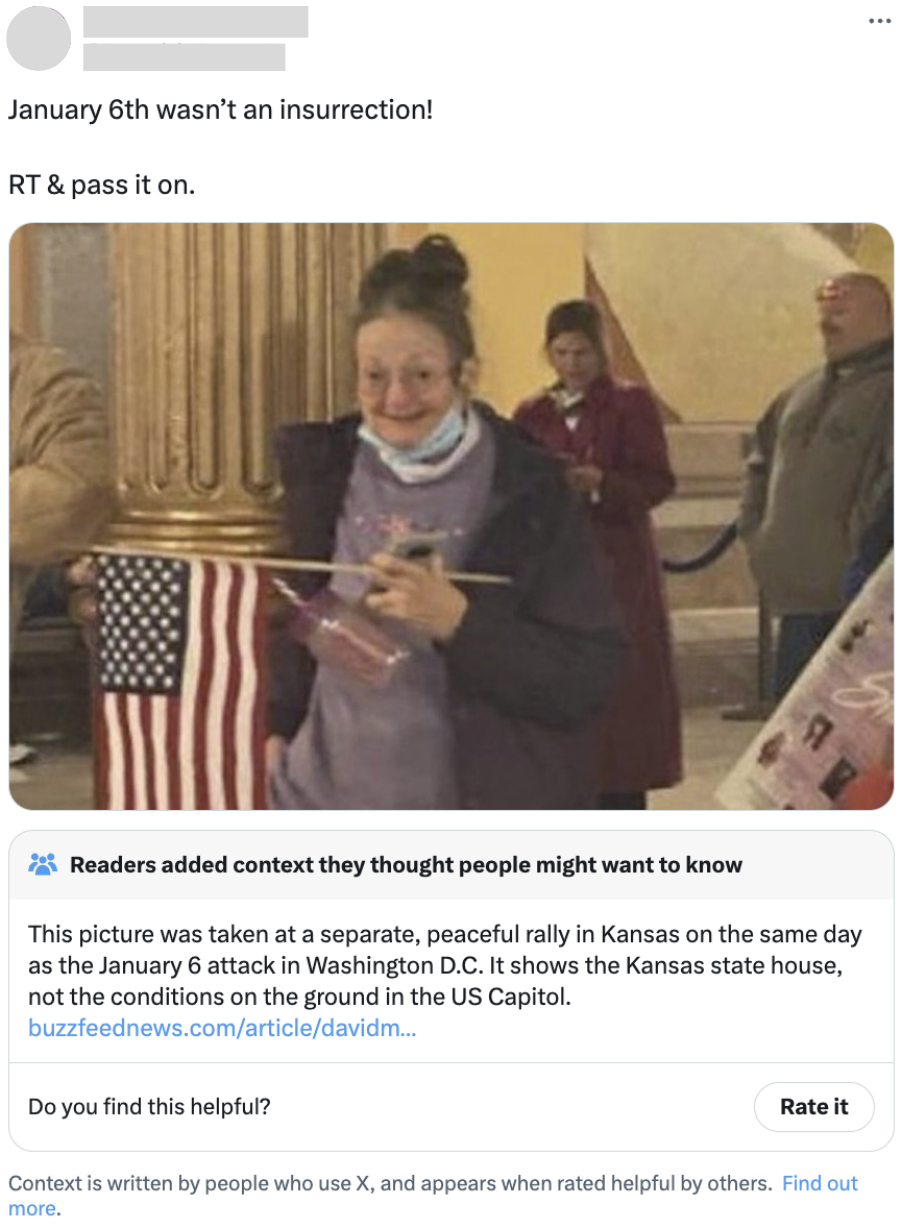}
    \caption{An example of a high-helpfulness response from Community Notes displayed on the corresponding tweet and visible to the public.}
    \label{supp:figure:example_helpful_note}
\end{figure}

\begin{figure}
    \centering
    \includegraphics[scale=0.9]{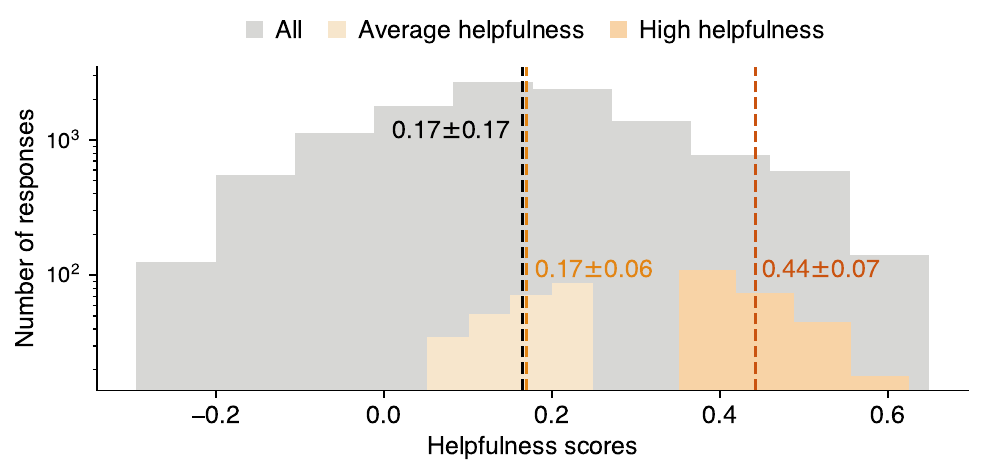}
    \caption{Distribution of helpfulness scores of laypeople's responses in X Community Notes. 
    All: All laypeople's responses in Community Notes.
    Average helpfulness: Laypeople's responses in Community Notes identified with average helpfulness and used in our study.
    High helpfulness: Laypeople's responses in Community Notes identified with high helpfulness and used in our study.
    For $x\pm y$, $x$: mean, $y$: standard deviation. Community Notes data are regularly updated; ours are up until February 12, 2023.}
    \label{supp:figure:dist_helpfulness_scores}
\end{figure}

\begin{figure}
    \centering
    \includegraphics[width=.95\textwidth]{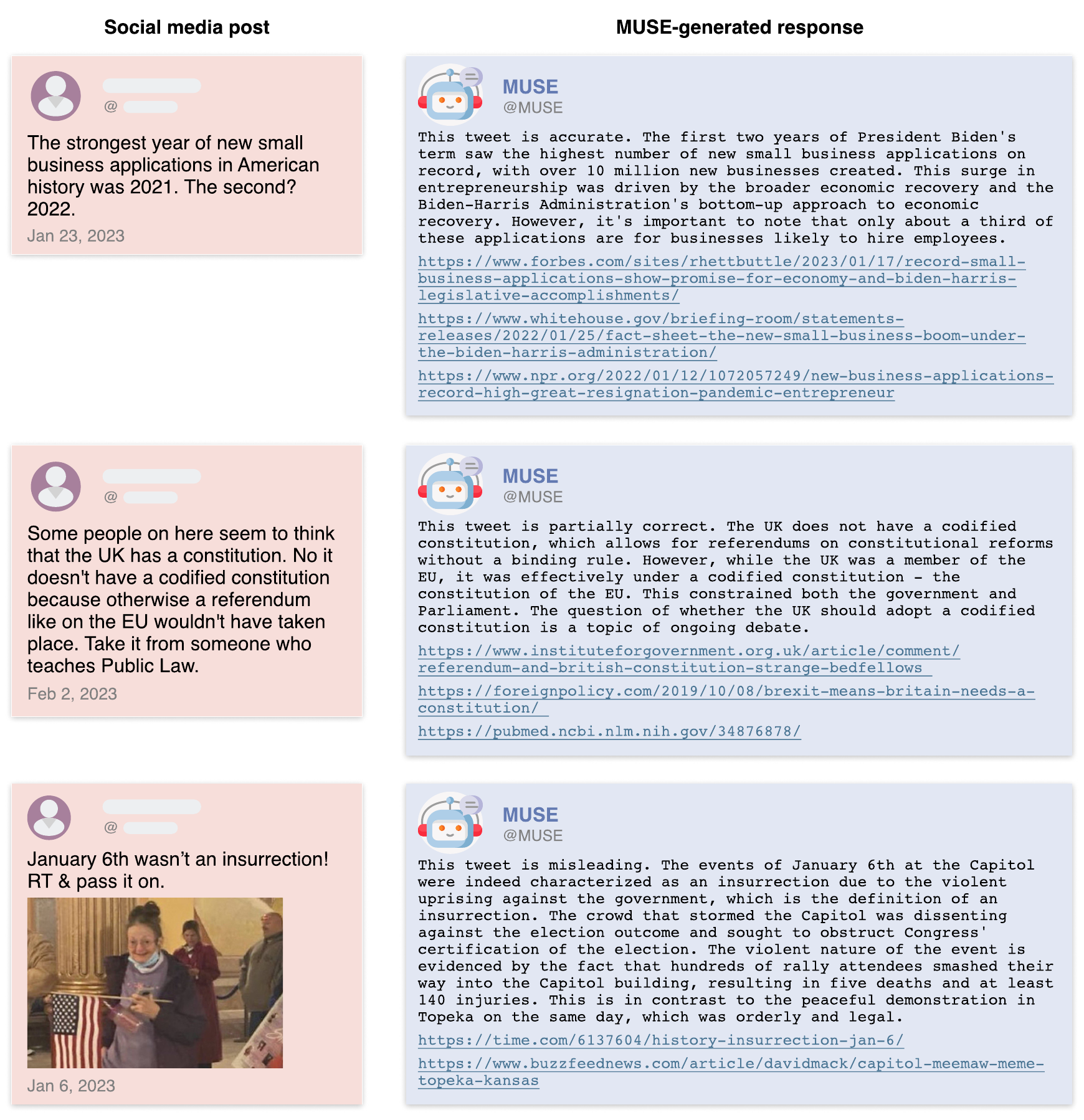}
    \caption{Examples of \oursystem-generated responses to accurate, partially accurate, and factually correct but misleading content on social media.}
    \label{supp:figure:example_muse_responses}
\end{figure}

\begin{figure}
    \centering
    \includegraphics[scale=1]{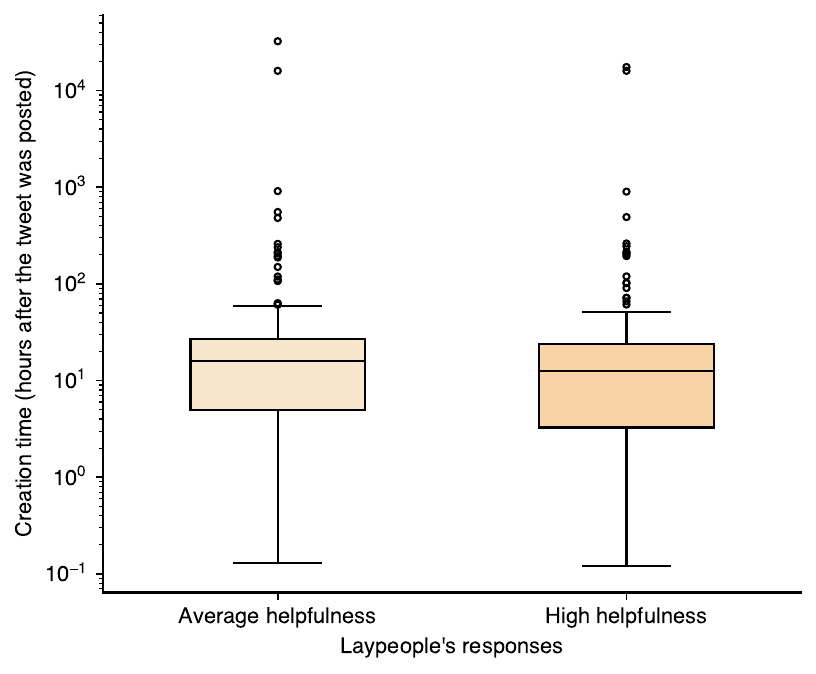}
    \caption{Distribution of creation times of laypeople's responses in X Community Notes used in our study. Median of the creation time of laypeople's average-helpfulness responses: 16 hours after the tweet was posted. Median of the creation time of laypeople's high-helpfulness responses: 13 hours after the tweet was posted.}
    \label{supp:figure:dist_note_times}
\end{figure}

\begin{figure}
    \centering
    \begin{subfigure}{0.26\textwidth}
        \includegraphics[width=\textwidth]{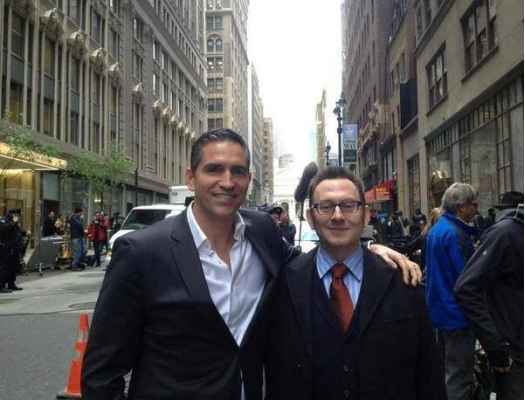}
        \caption{}
    \end{subfigure}
    \begin{subfigure}{0.2\textwidth}
        \includegraphics[width=\textwidth]{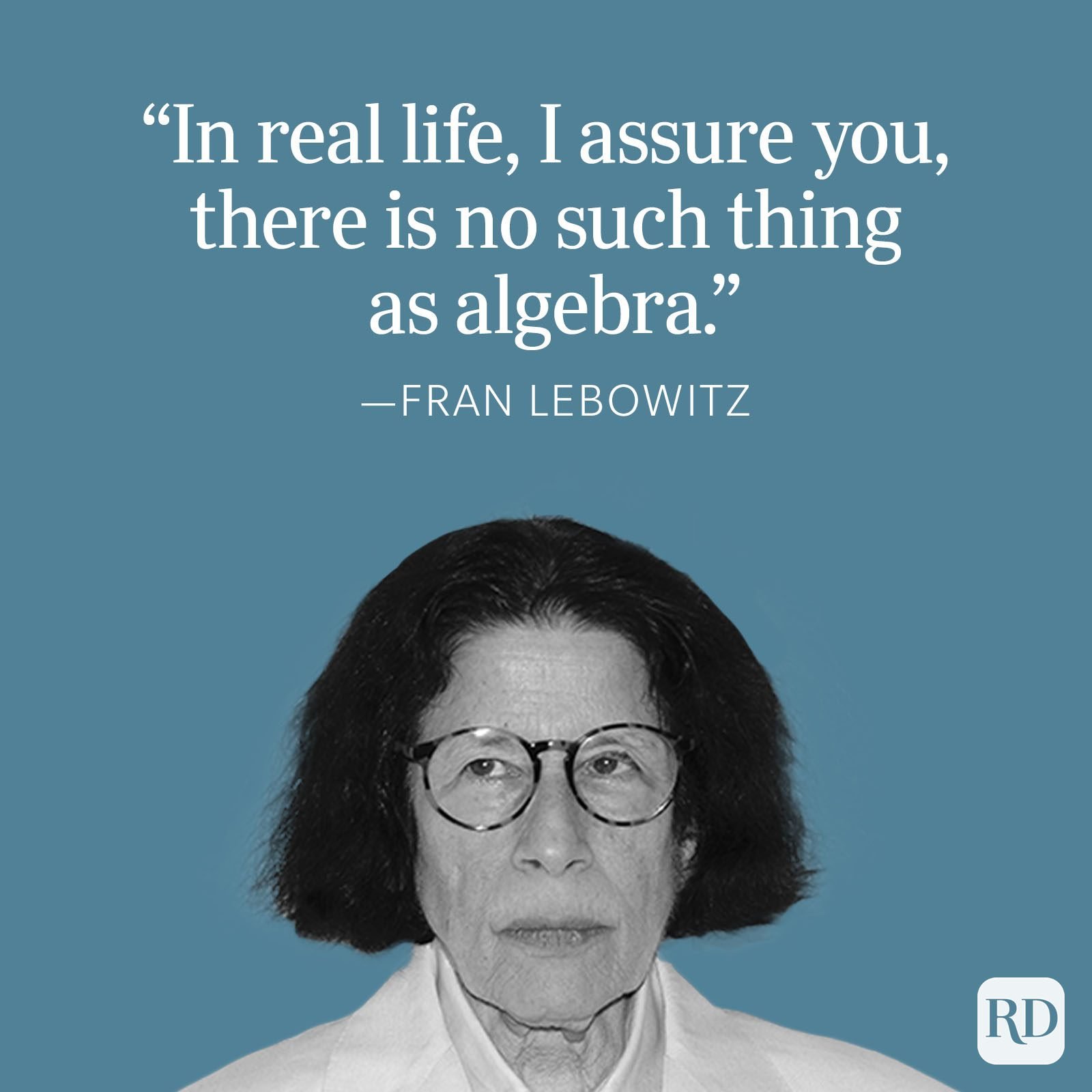}
        \caption{}
    \end{subfigure}
    \begin{subfigure}{0.2\textwidth}
        \includegraphics[width=\textwidth]{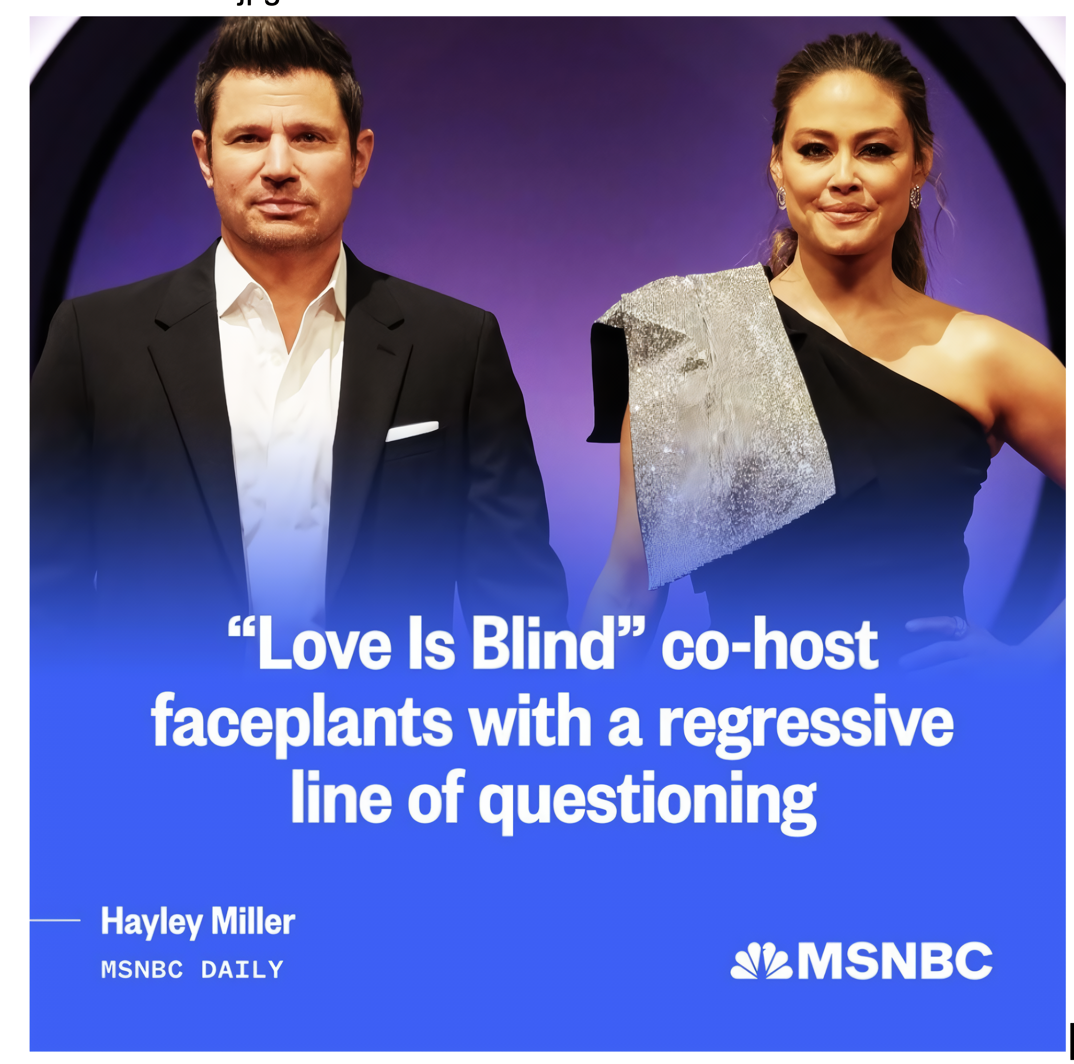}
        \caption{}
    \end{subfigure}
    \begin{subfigure}{0.31\textwidth}
        \includegraphics[width=\textwidth]{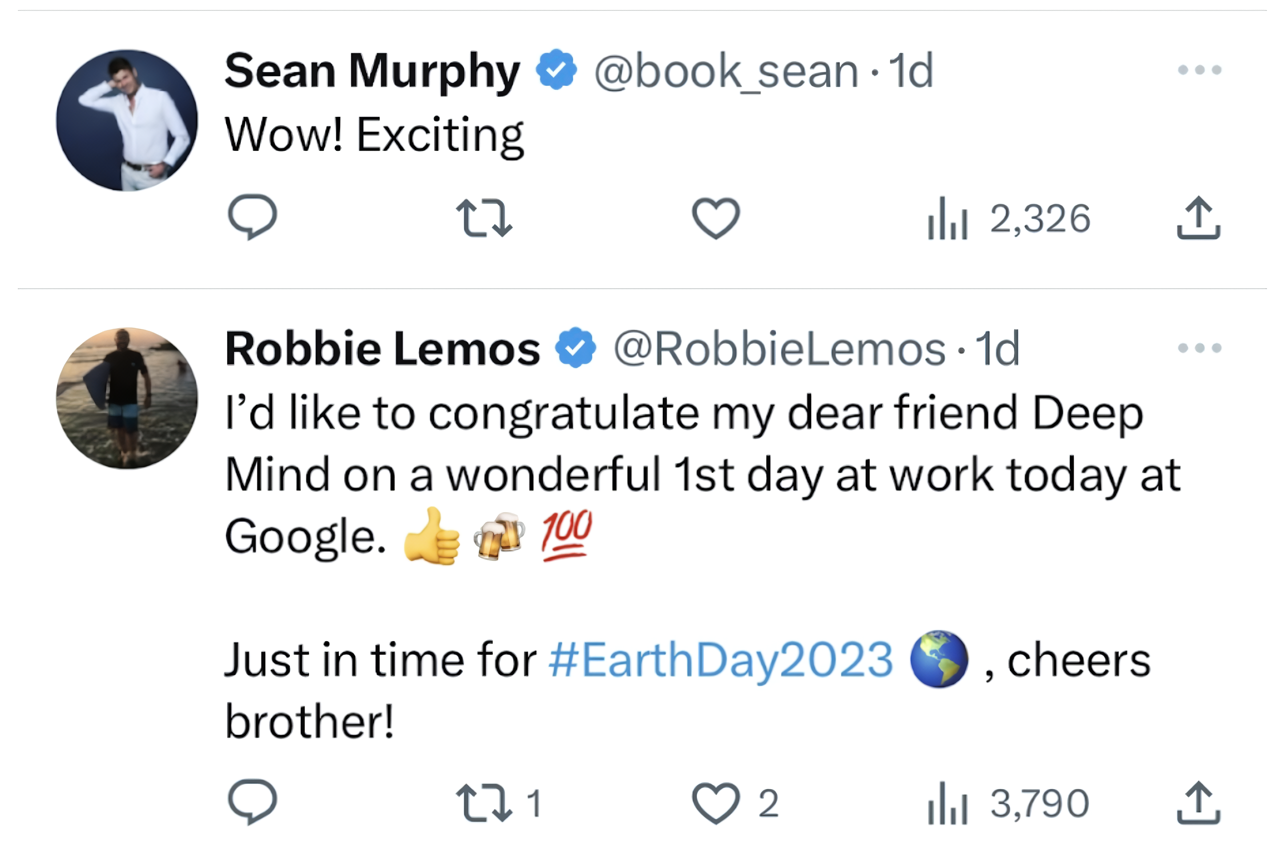}
        \caption{}
    \end{subfigure}

    \begin{subfigure}{0.49\textwidth}
        \includegraphics[width=0.85\textwidth]{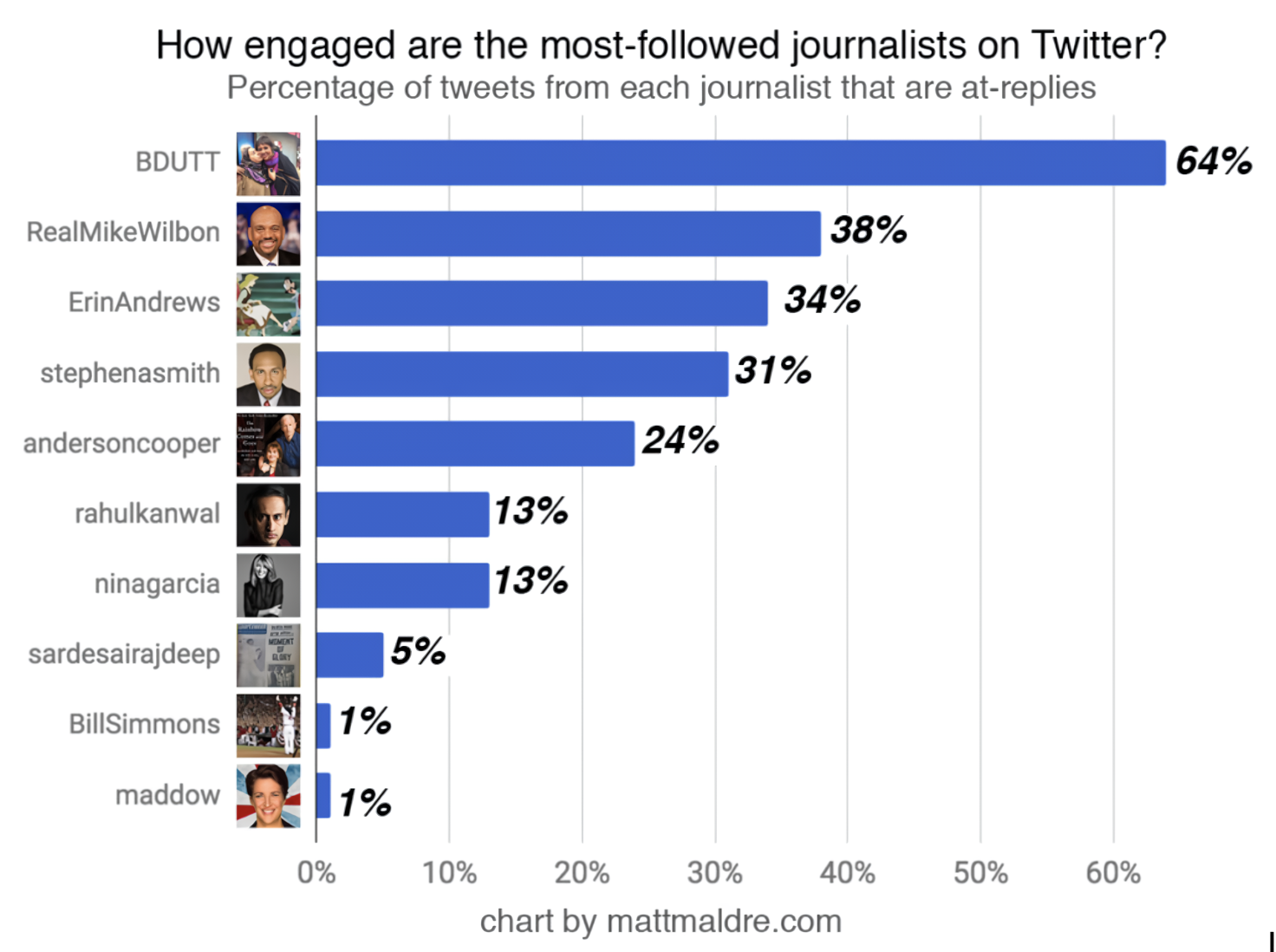}
        \caption{}
    \end{subfigure}\hfill
    \begin{subfigure}{0.49\textwidth}
        \includegraphics[width=\textwidth]{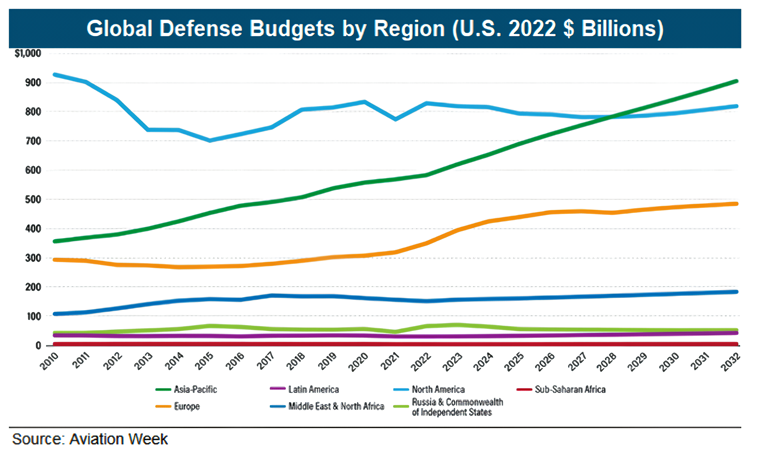}
        \caption{}
    \end{subfigure}

    \begin{subfigure}{0.49\textwidth}\centering
        \includegraphics[width=0.6\textwidth]{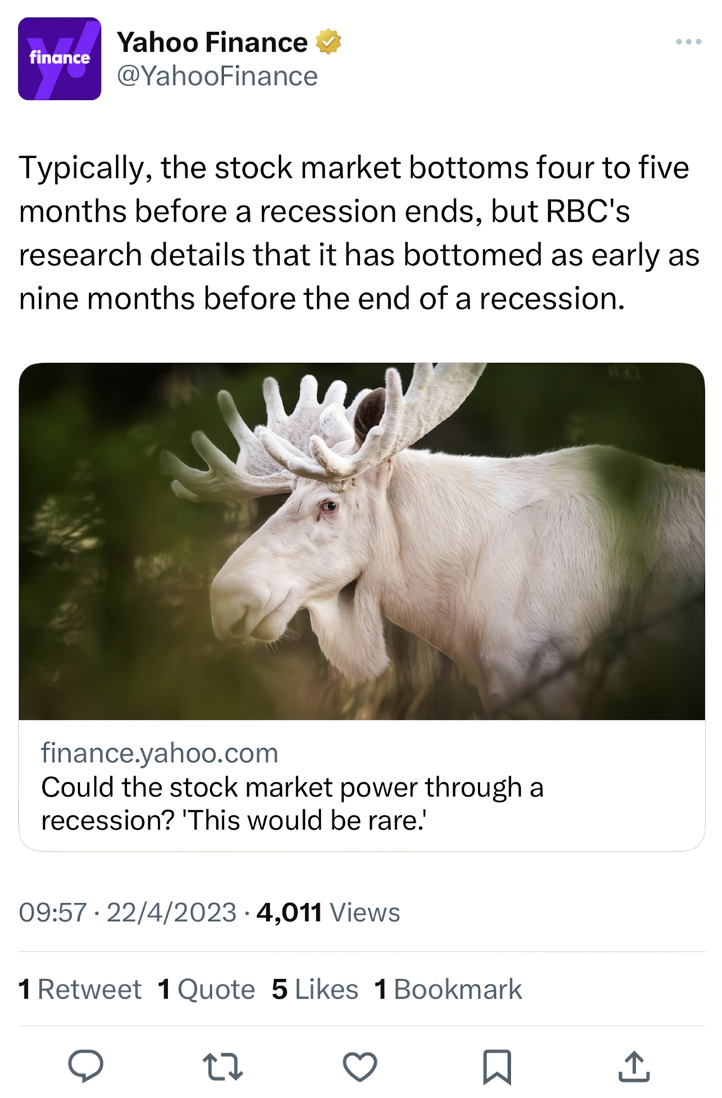}
        \caption{}
    \end{subfigure}\hfill
    \begin{subfigure}{0.49\textwidth}
        \includegraphics[width=\textwidth]{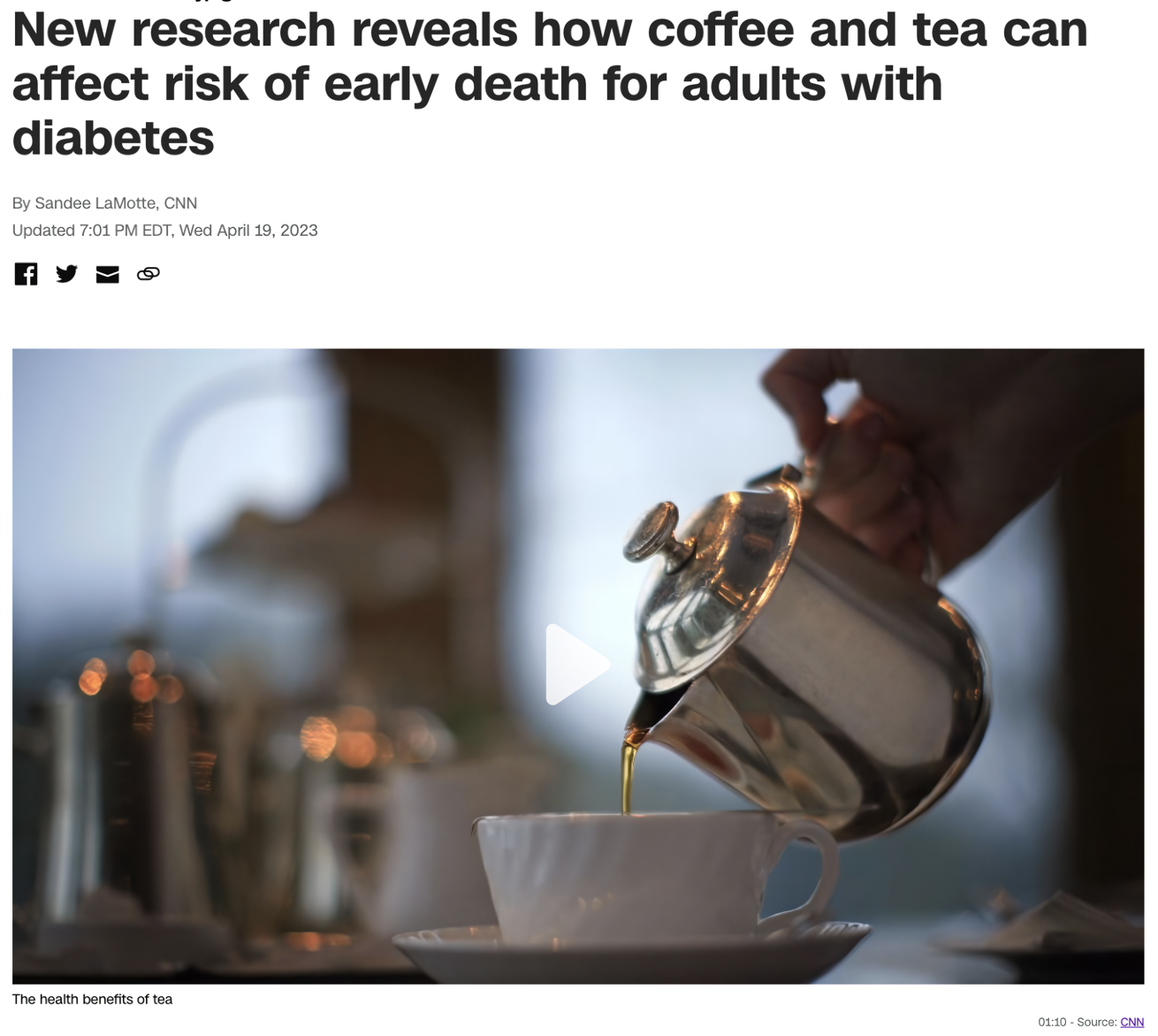}
        \caption{}
    \end{subfigure}
    
    \caption{Images as examples used for informative image captioning with in-context learning. These example images do not appear in the dataset we used.}
    \label{supp:figure:example_images}
\end{figure}

\begin{figure}
    \centering
    \includegraphics[width=.9\textwidth]{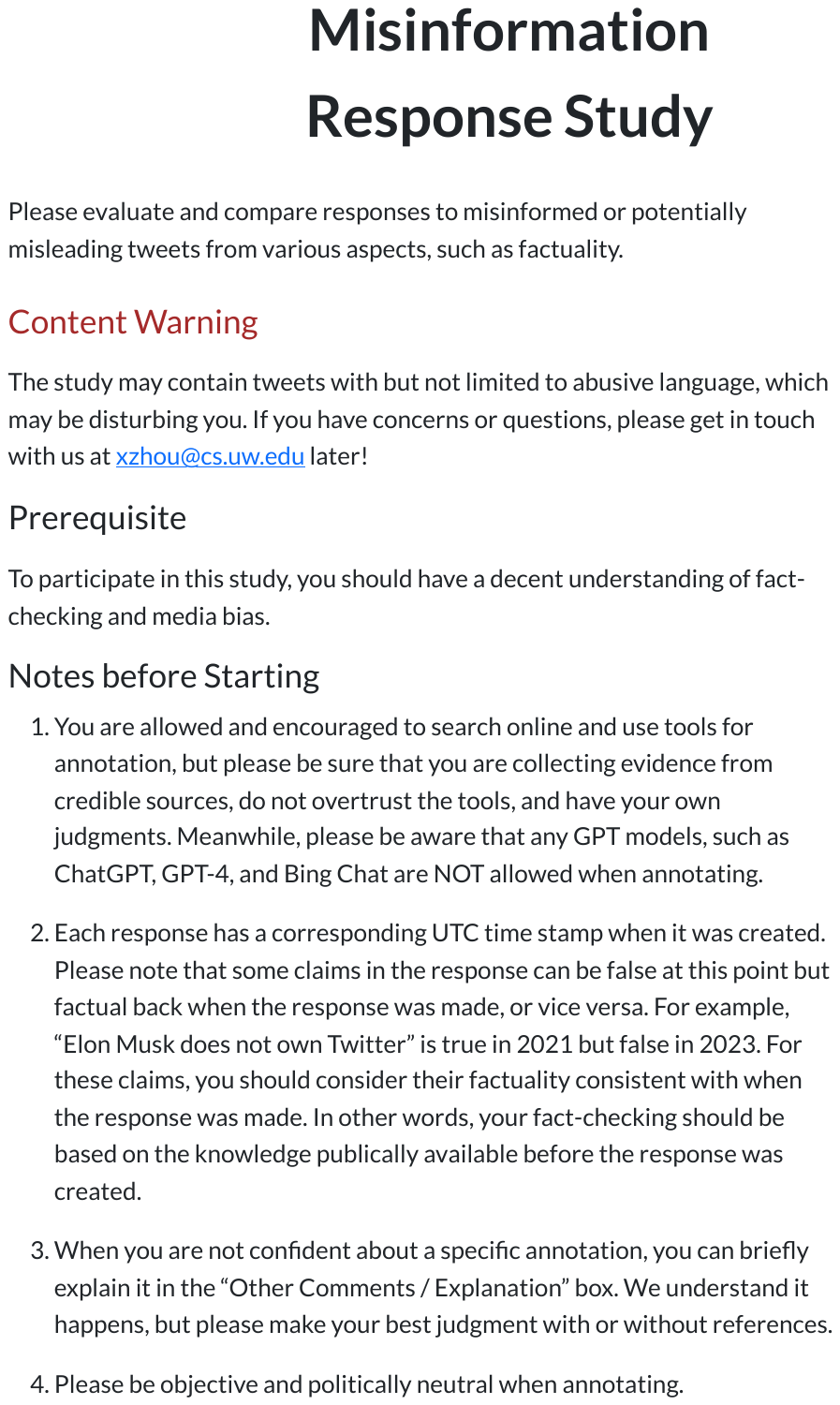}
    \caption{Annotation instructions (page 1/7, continued on the next page).}
    \label{supp:figure:annotation_instructions_1}
\end{figure}

\begin{figure}
    \centering
    \includegraphics[width=.9\textwidth]{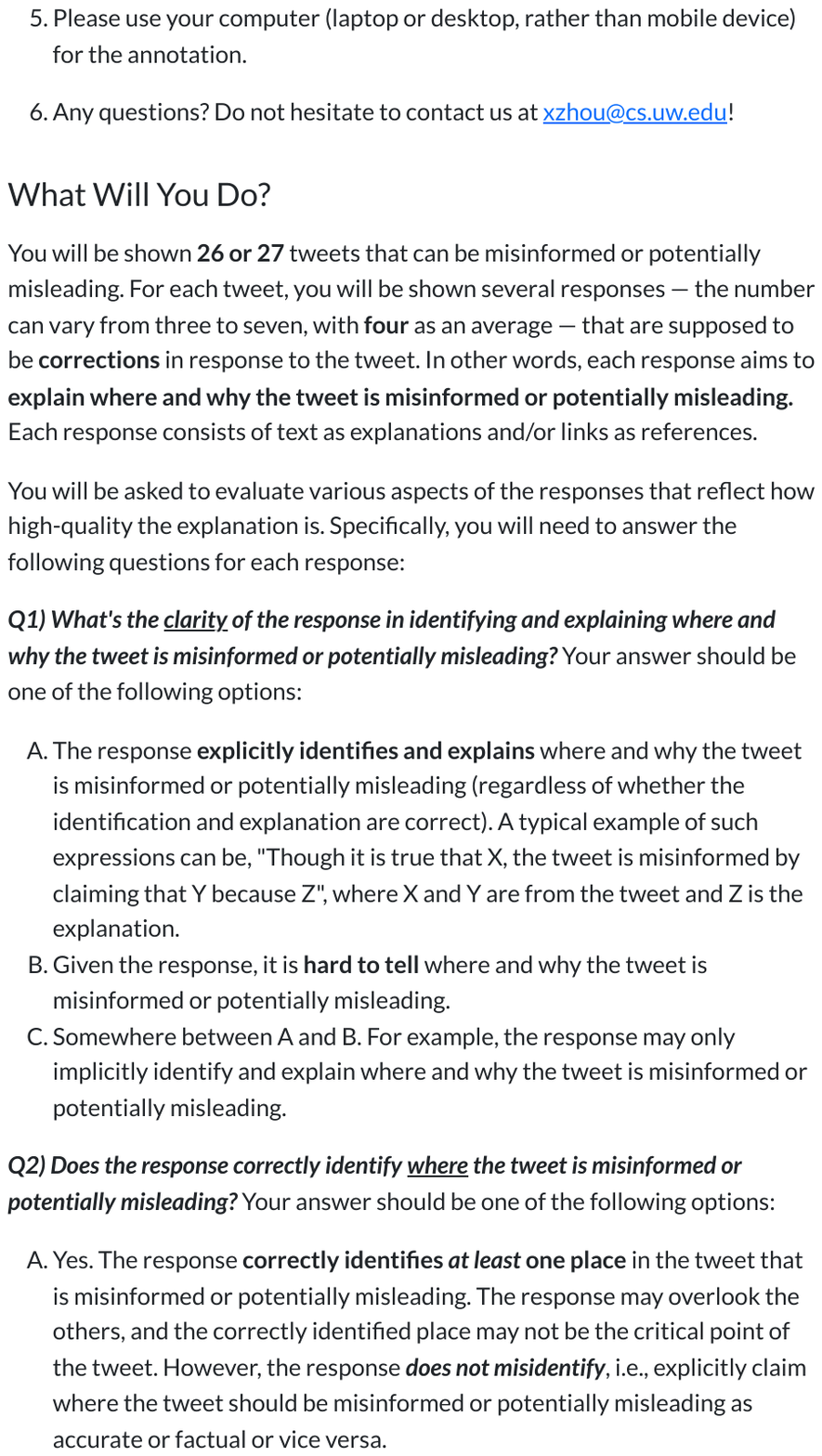}
    \caption{Annotation instructions (page 2/7, continued on the next page).}
    \label{supp:figure:annotation_instructions_2}
\end{figure}

\begin{figure}
    \centering
    \includegraphics[width=.9\textwidth]{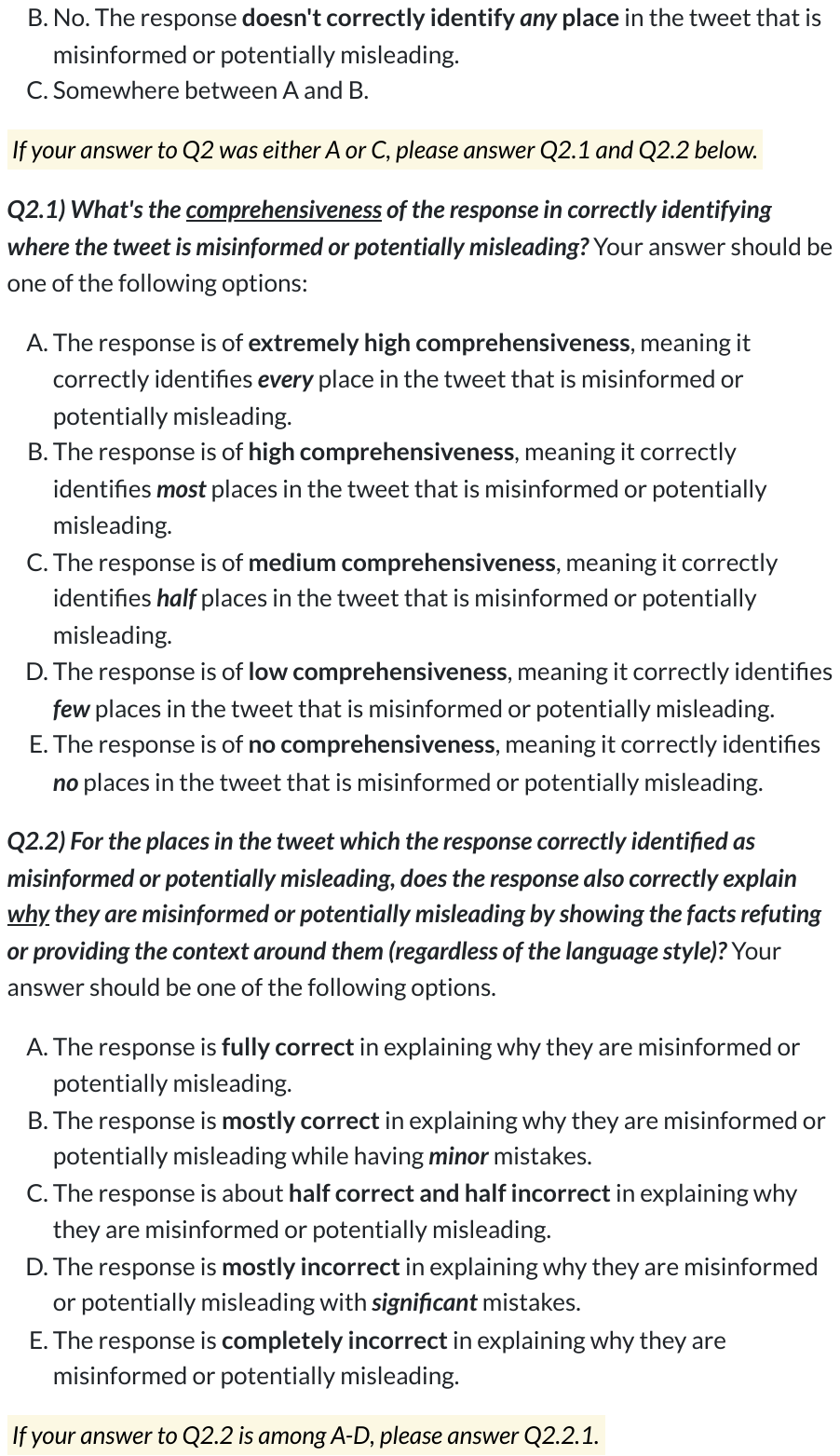}
    \caption{Annotation instructions (page 3/7, continued on the next page).}
    \label{supp:figure:annotation_instructions_3}
\end{figure}

\begin{figure}
    \centering
    \includegraphics[width=.9\textwidth]{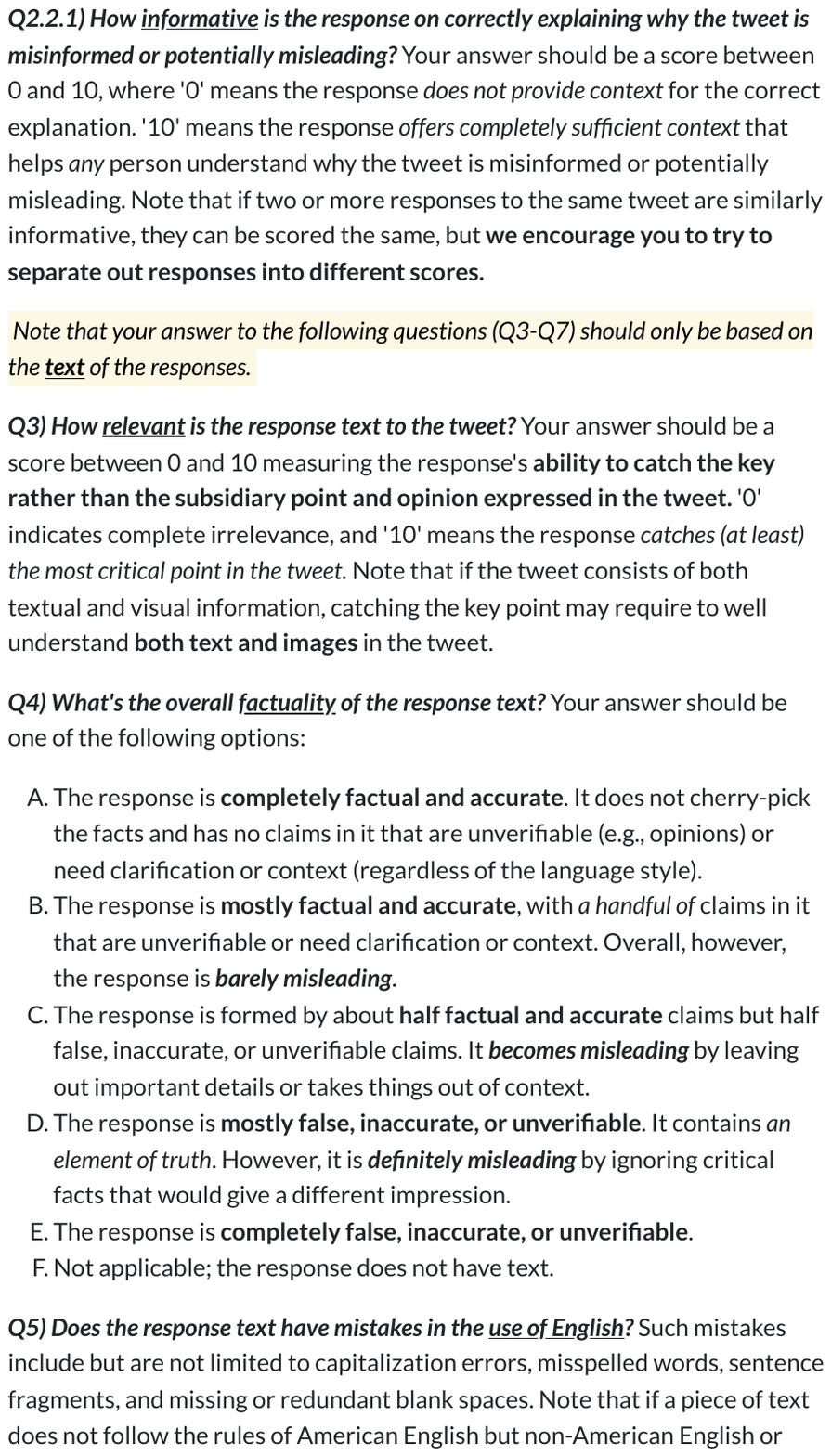}
    \caption{Annotation instructions (page 4/7, continued on the next page).}
    \label{supp:figure:annotation_instructions_4}
\end{figure}

\begin{figure}
    \centering
    \includegraphics[width=.9\textwidth]{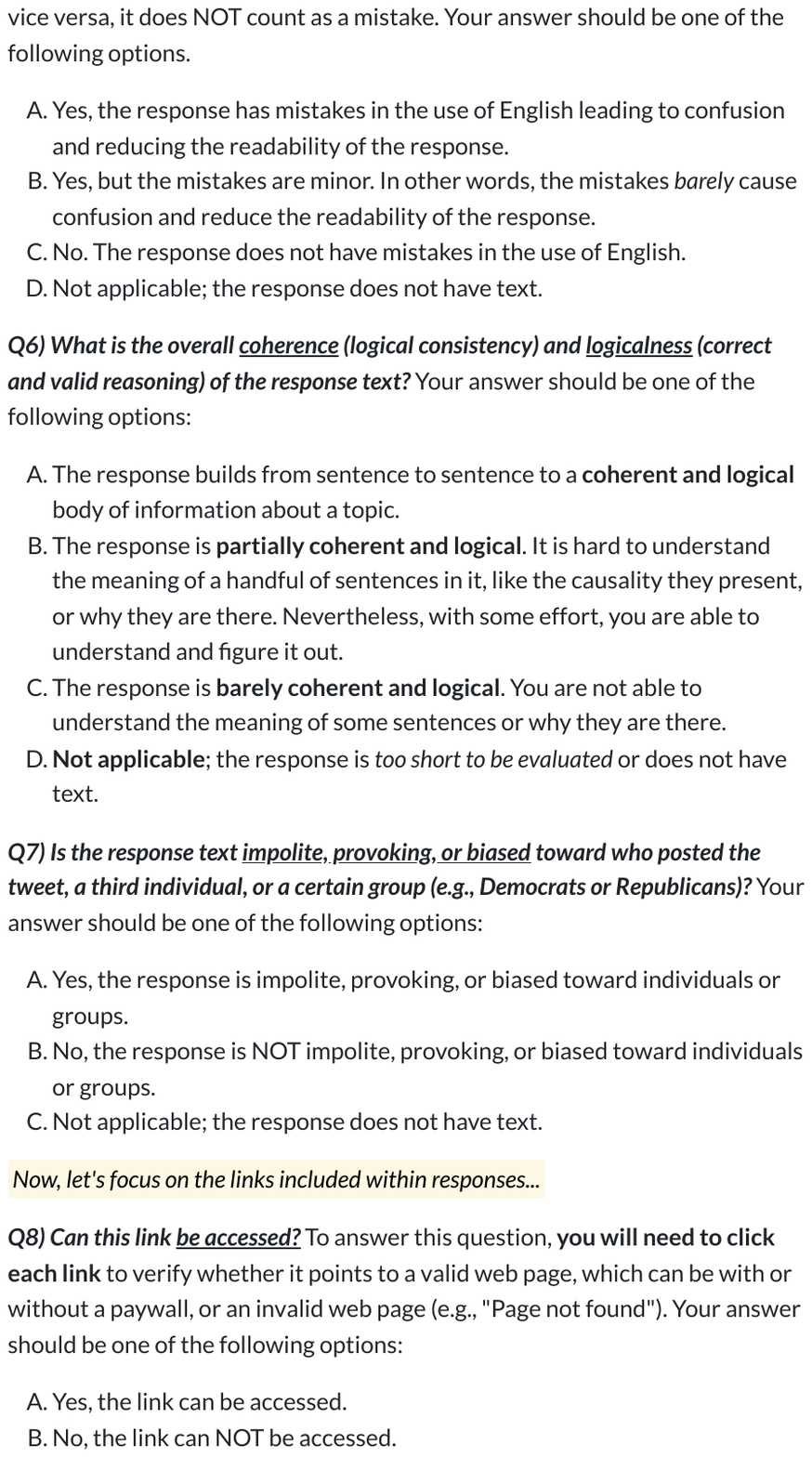}
    \caption{Annotation instructions (page 5/7, continued on the next page).}
    \label{supp:figure:annotation_instructions_5}
\end{figure}

\begin{figure}
    \centering
    \includegraphics[width=.9\textwidth]{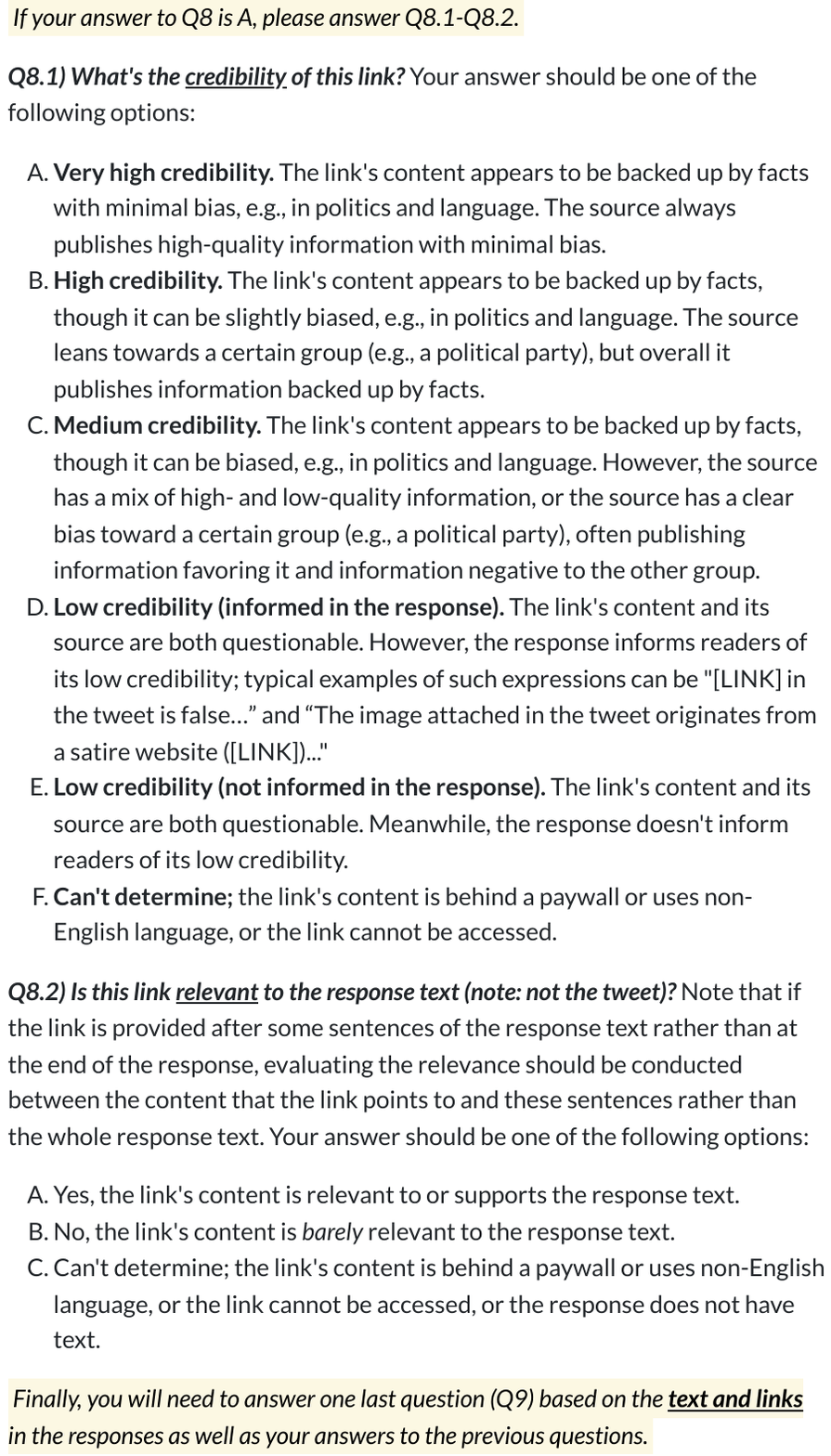}
    \caption{Annotation instructions (page 6/7, continued on the next page).}
    \label{supp:figure:annotation_instructions_6}
\end{figure}

\begin{figure}
    \centering
    \includegraphics[width=.9\textwidth]{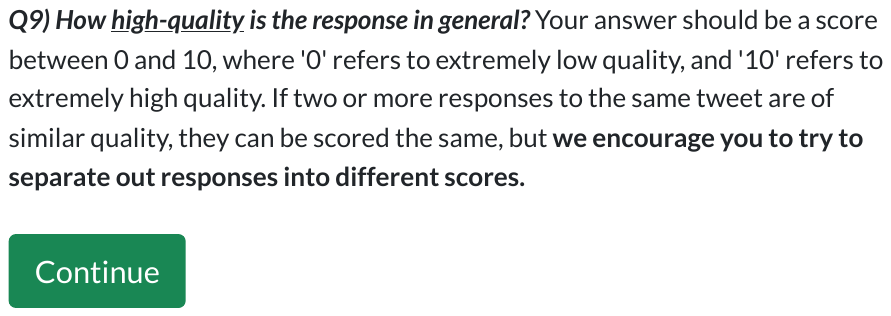}
    \caption{Annotation instructions (page 7/7).}
    \label{supp:figure:annotation_instructions_7}
\end{figure}

\begin{figure}
    \centering
    \includegraphics[width=\textwidth]{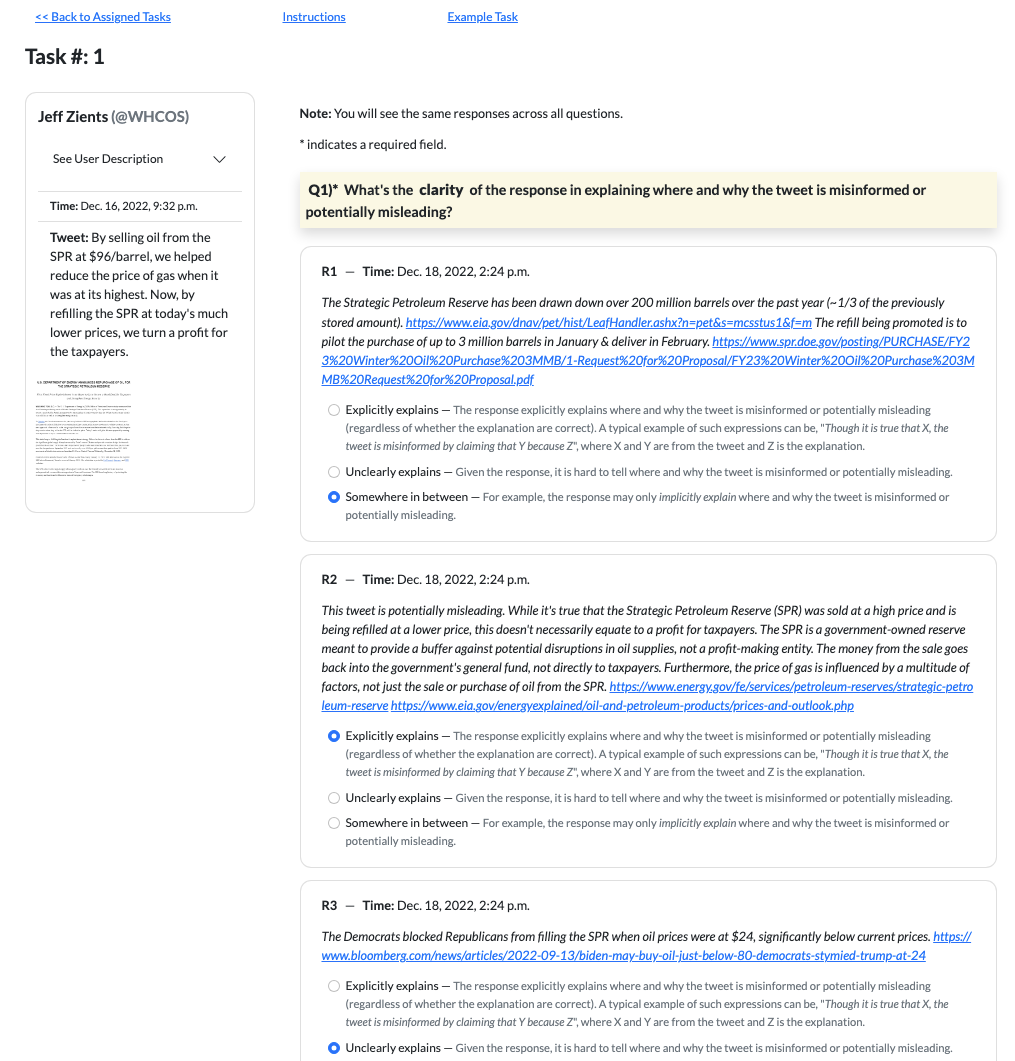}
    \caption{Annotation task page.}
    \label{supp:figure:annotation_task_page}
\end{figure}

\begin{figure}
    \centering
    \includegraphics[width=0.98\textwidth]{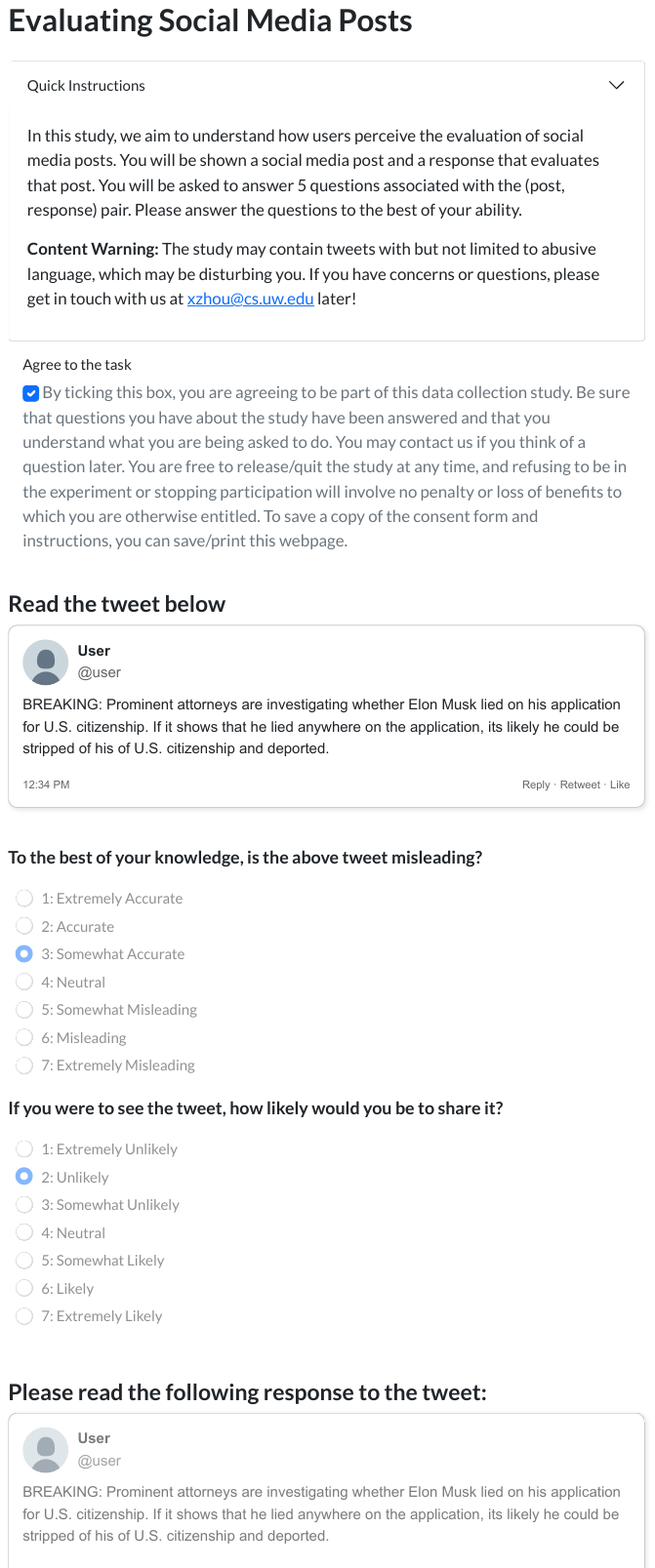}
    \caption{Study interface to measure end user perceptions of corrections (page 1/2, continued on the next page).}
    \label{supp:figure:user-study-interface-1}
\end{figure}

\begin{figure}
    \centering
    \includegraphics[width=0.98\textwidth]{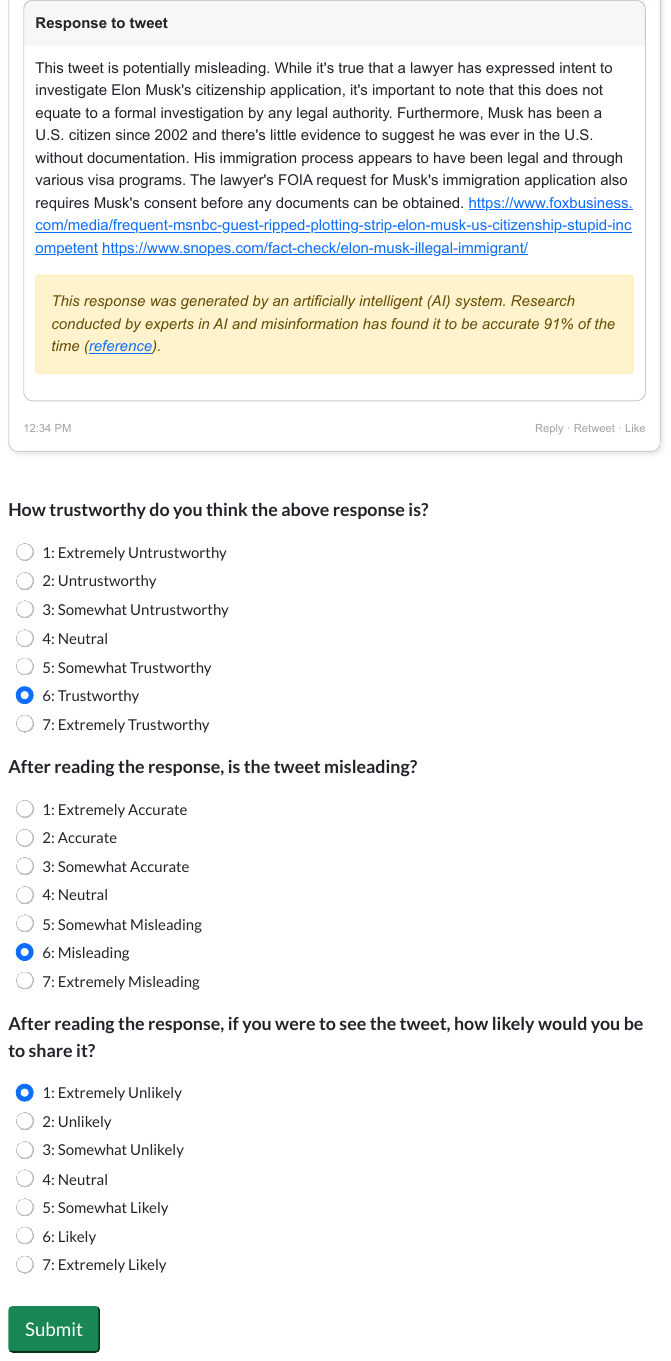}
    \caption{Study interface to measure end user perceptions of corrections (page 2/2).}
    \label{supp:figure:user-study-interface-2}
\end{figure}

\begin{figure}
    \centering
    \includegraphics[width=.97\textwidth]{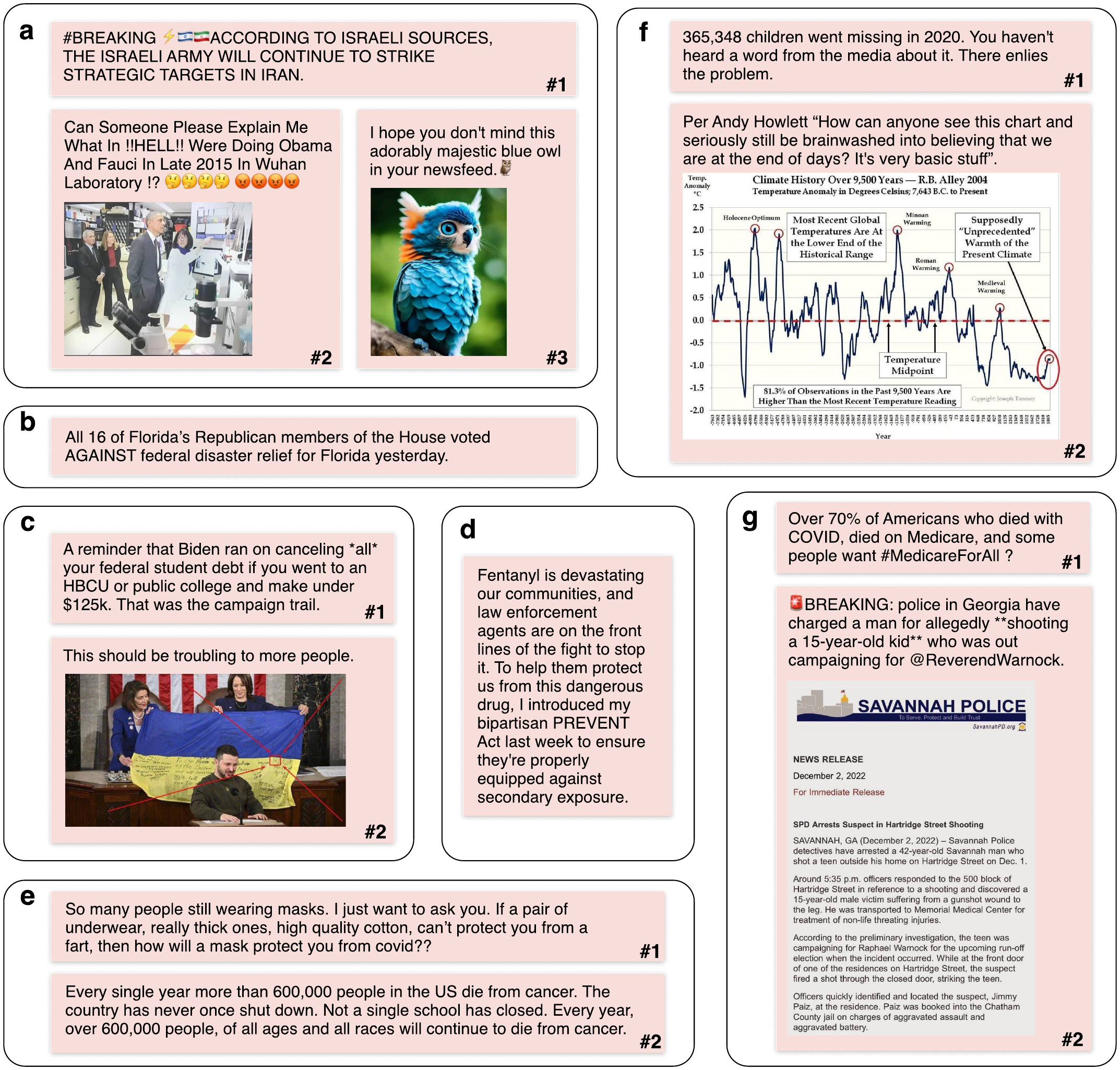}
    \caption{Examples of social media content that uses various tactics to make it or part of it false or misleading. These tactics include
    \textbf{a}: \textit{Fabricating} a news event (\#1), a story with an authentic image (\#2, also related to conspiracy theory), and an image with AI (\#3).
    \textbf{b}: \textit{Lacking context} regarding the motivation behind behavior. Here, it is the consideration that the bill contained no funding for Florida, without which could mislead readers.
    \textbf{c}: \textit{Misinterpreting or misrepresenting} a plan, where Biden did not promise to cancel all the debt (\#1), or symbol, which is not Nazi SS runes but a shorthand for the 46th Separate Airmobile Brigade (\#2).
    \textbf{d}: \textit{Using loaded language}.
    \textbf{e}: \textit{Improperly analogizing or equating} masks blocking respiratory droplets and underwear blocking gas molecules (\#1), and COVID-19 and cancer (\#2).
    \textbf{f}: \textit{Presenting false, partial, or biased data}. 365348 includes multiple reports for the same child and runaways who may not be considered missing because of their guardians know their whereabouts (\#1). The data ends in 1885 and is from a specific high elevation site in Greenland, not global temperatures (\#2).
    \textbf{g}: \textit{Implying false or oversimplified causation}. Many COVID-19 deaths occurred among Medicare beneficiaries because Medicare primarily serves the groups who are at greater risk of adverse outcomes from COVID-19 (\#1). The boy was shot when---but not because---he was campaigning for Raphael Warnock (\#2). Note that one content may apply more than one tactic intentionally or accidentally.
    }
    \label{supp:figure:example_misinfo_tactic}
\end{figure}

\begin{figure}

    \centering

    \begin{subfigure}{\textwidth}\centering
        \includegraphics[scale=0.9]{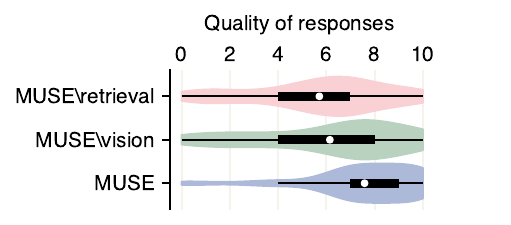}
        \caption{Overall quality of responses.}
    \end{subfigure}

    \begin{subfigure}{\textwidth}\centering
        \includegraphics[scale=0.62]{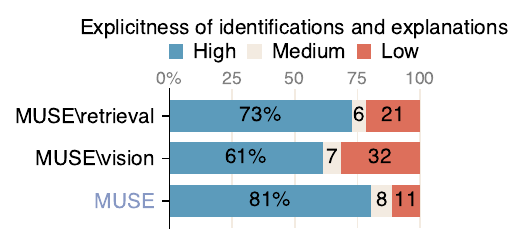}
        \includegraphics[scale=0.62]{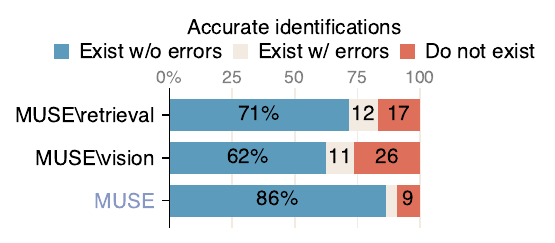}
        \includegraphics[scale=0.62]{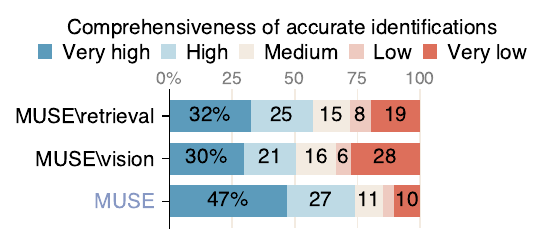}
        \includegraphics[scale=0.62]{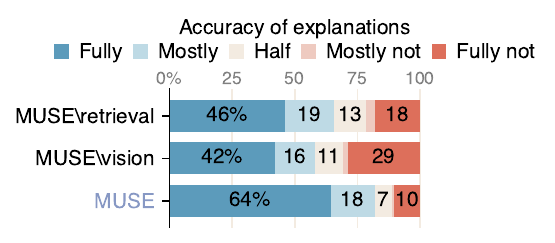}
        \includegraphics[scale=0.62]{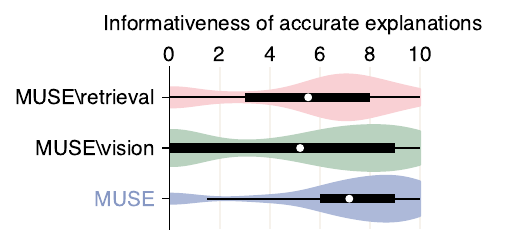}
        \caption{Quality of responses in identifying and explaining (in)accuracies.}
    \end{subfigure}
    
    \begin{subfigure}{\textwidth}\centering
        \includegraphics[scale=0.62]{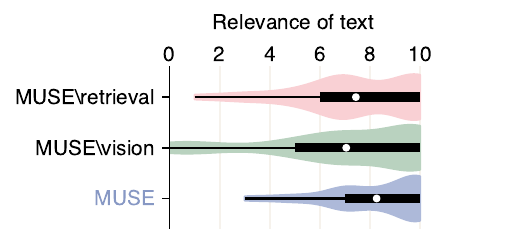}
        \includegraphics[scale=0.62]{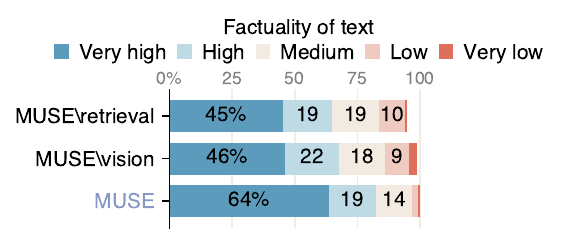}
        \includegraphics[scale=0.62]{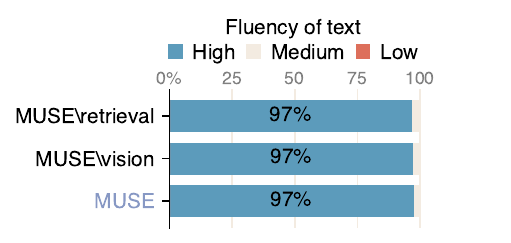}
        \includegraphics[scale=0.62]{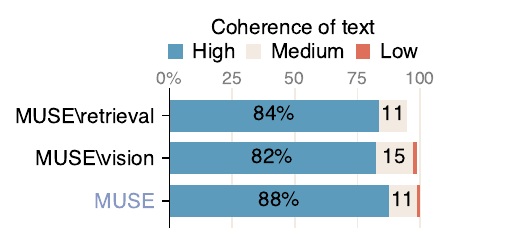}
        \includegraphics[scale=0.62]{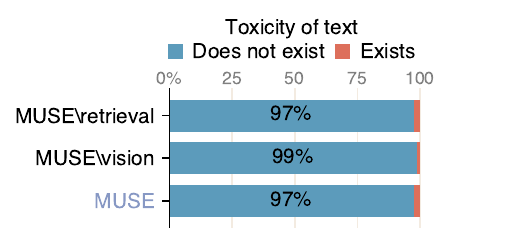}
        \caption{Quality of responses in generated text.}
    \end{subfigure}
    
    \begin{subfigure}{\textwidth}\centering
        \includegraphics[scale=0.62]{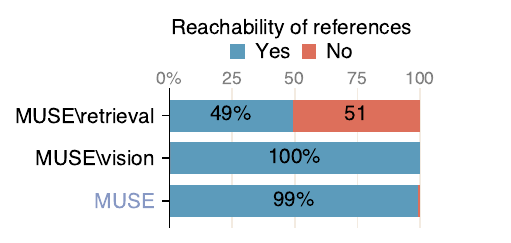}
        \includegraphics[scale=0.62]{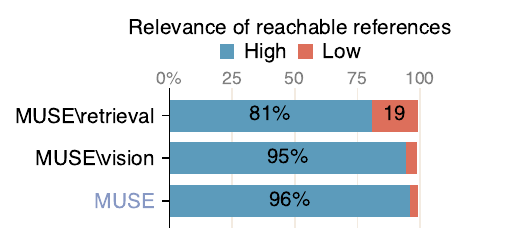}
        \includegraphics[scale=0.62]{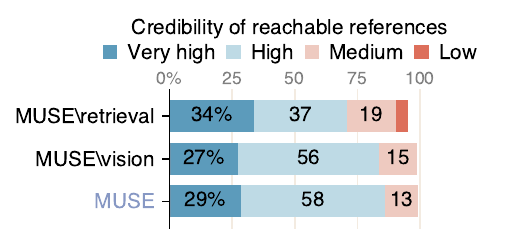}
        \caption{Quality of responses in references.}
    \end{subfigure}
    
    \caption{Impact of retrieval and vision on \oursystem's performance. Here, \oursystem~and \oursystem$\backslash$vision responded to tweets by only retrieving web pages published thirty minutes before the creation time of the corresponding laypeople's high-helpfulness response.}
    \label{supp:figure:impact_retrieval_and_vision}
    
\end{figure}

\begin{figure}
    \centering
    \includegraphics[scale=1]{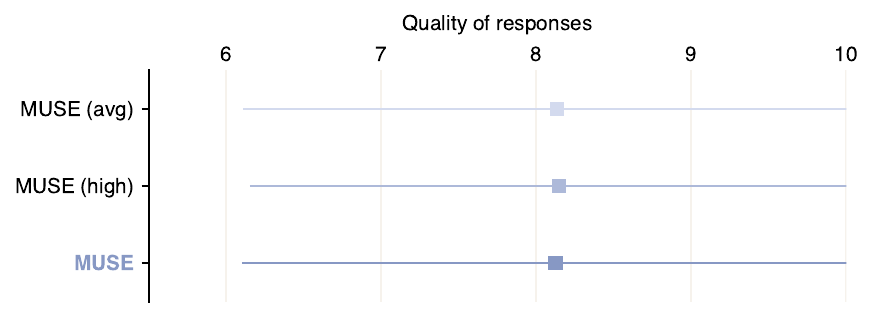}
    \caption{Impact of starting times of responding to tweets on \oursystem's performance. 
    The simulated starting time for \oursystem~(avg): Thirty minutes before the corresponding laypeople's average-helpfulness responses was created (median: 16 hours after the corresponding tweet was posted; Supplementary Fig.~\ref{supp:figure:dist_note_times}). 
    The simulated starting time for \oursystem~(high): Thirty minutes before the corresponding laypeople's high-helpfulness responses was created (median: 13 hours after the corresponding tweet was posted; Supplementary Fig.~\ref{supp:figure:dist_note_times}). 
    The simulated starting time for \oursystem: The post time of the corresponding tweet (i.e., 0 hours after the corresponding tweet was posted).}
    \label{supp:figure:impact_time_correction}
\end{figure}

\begin{figure}
    \centering
    \includegraphics[width=.9\textwidth]{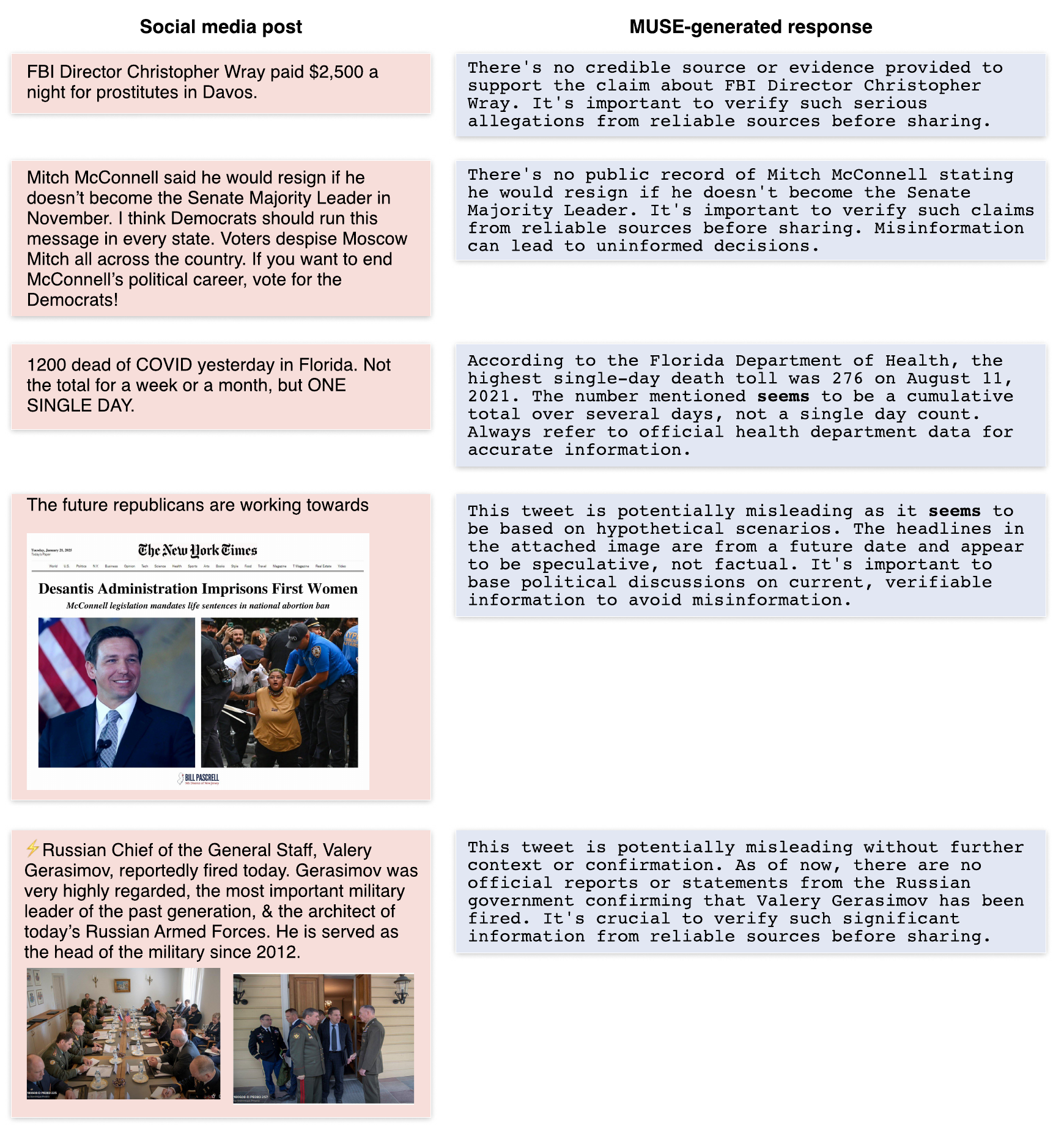}
    \caption{Examples of \oursystem-generated responses to social media posts that lack evidence to verify.}
    \label{supp:figure:example_no_evidence}
\end{figure}

\begin{figure}
    \centering
    \includegraphics[width=\textwidth]{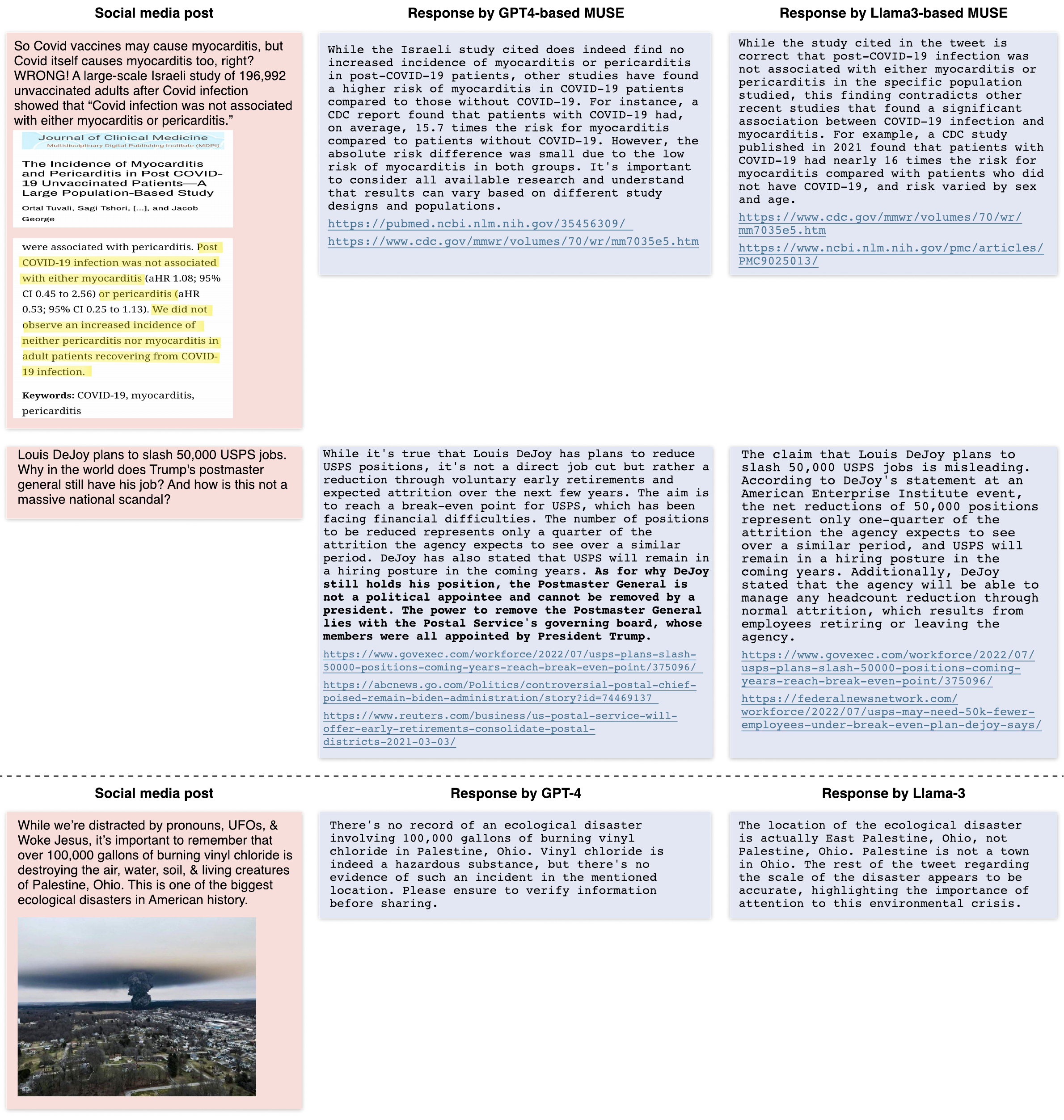}
    \caption{Comparison between responses generated by GPT-4, Llama-3, and their based \oursystem. We observed that GPT4-based \oursystem{} and Llama3-based \oursystem{} frequently generate similar responses to social media content (see the top example). GPT4-based \oursystem{} occasionally generates more comprehensive corrections than Llama3-based \oursystem{} (see the middle example). Note that the knowledge cutoff for Llama3 is December 2023, whereas that of GPT-4 is September 2021. In other words, Llama-3's training data include more recent events and knowledge than GPT-4's (see the bottom example, where the post was posted in February 2023, and Llama-3 more accurately identifies its falsehood than GPT-4).}
    \label{supp:figure:impact_foundation_llm}
\end{figure}

List of supplementary materials:\\
Table~\ref{supp:table:mbfc}\\
Figures~\ref{supp:figure:timeliness_notes}--\ref{supp:figure:impact_foundation_llm}

\newpage
\begin{small}
\begin{longtable}{p{3.2cm}p{11.4cm}r}
\caption{Definitions and statistics of factuality and bias categories offered by Media Bias/Fact Check (\href{https://mediabiasfactcheck.com/}{mediabiasfactcheck.com}).}
\label{supp:table:mbfc}\\
\toprule
Category & Definition & \makecell[rb]{\# of sources} \\ \midrule
\textbf{Factuality:} &  &  \\
- Very high: & The source is consistently factual, relies on credible information, promptly corrects errors, and has never failed any fact checks in news reporting or opinion pieces. & 118 \\
- High: & The source is mostly factual and uses mostly credible, low-biased, or high-factual sources. It corrects errors quickly and has failed only one news fact check and up to two op-ed fact checks. & 2,313 \\
- Mostly factual: & The source is generally accurate but may have a few uncorrected fact-check failures. It can fail up to three op-ed fact checks, especially if it is a low-volume site. While it may use biased sources occasionally, it mostly links to factual content. It is usually pro-science but may sometimes use misleading wording or offer alternative viewpoints. It is reasonably transparent and trustworthy most of the time, but caution is advised. & 326 \\
- Mixed: & The source may rely on improper sourcing or link to other biased or mixed-factual sources. It often has multiple failed fact checks and does not correct false information or lacks transparency, including the absence of a disclosed mission statement or ownership details. 
    Sources rejecting established scientific consensus on issues such as climate change or vaccines will receive this rating or lower. 
    Sources identified as overt propaganda or designated as hate groups by reputable third-party evaluators will receive this rating or lower due to their inherent bias and potential spread of misleading information. & 1,437 \\
- Low: & The source is often unreliable and should be fact-checked for fake news, conspiracy theories, and propaganda. & 677 \\
- Very low: & The source is almost always unreliable and should always be fact-checked for intentional misinformation. & 252 \\
\multicolumn{3}{r}{\textbf{Total: 5,123}} \\ \midrule

\textbf{Bias:} & \textbf{} & \textbf{} \\
- Least biased: & The source has minimal bias and uses very few loaded words (i.e., wording that attempts to influence an audience by using an appeal to emotion or stereotypes). It is factual and usually sourced. & 1,054 \\
- Left-center: & The source has a slight to moderate liberal bias. It often publishes factual information that utilizes loaded words to favor liberal causes. It is generally trustworthy for information but may require further investigation. & 850 \\
- Right-center: & Similar to the definition of left-center bias but replacing liberal with conservative. & 492 \\
- (Extremely) left: & The source is moderately to strongly biased toward liberal causes through story selection or political affiliation. It may utilize strong loaded words, publish misleading reports, and omit reporting of information that may damage liberal causes. It may be untrustworthy. & 402 \\
- (Extremely) right: & Similar to the definition of (extremely) left bias but replacing liberal with conservative. & 314 \\
- Pro-science: & The source consists of legitimate science or is evidence-based through the use of credible scientific sourcing. Legitimate science follows the scientific method, is unbiased, and does not use emotional words. The source also respects the consensus of experts in the given scientific field and strives to publish peer-reviewed science. It may have a slight political bias but adheres to scientific principles. & 189 \\
\makecell[lt]{- Conspiracy-\\pseudoscience:} & The source may publish unverifiable information not always supported by evidence. It may be untrustworthy for credible or verifiable information, so fact-checking and further investigation are recommended on a per-article basis when obtaining information from it. & 433 \\
- Questionable: & The source exhibits one or more of the following: extreme bias, consistent promotion of propaganda or conspiracies, poor or no sourcing of credible information, a complete lack of transparency, or fake news (i.e., the deliberate attempt to publish hoaxes or disinformation for profit or influence). It may be very untrustworthy and should be fact-checked on a per-article basis. & 1,390 \\
- Satire: & The source exclusively uses humor, irony, exaggeration, or ridicule to expose and criticize people's stupidity or vices, particularly in the context of contemporary politics and other topical issues. It does not attempt to deceive. & 148 \\
\multicolumn{3}{r}{\textbf{Total: 5,272}} \\
 \bottomrule
\end{longtable}
\end{small}

% Introduction

% Approach

% Experiments

\begin{figure}
\begin{footnotesize}
\begin{Verbatim} [breaklines,numbers=left,xleftmargin=5mm]
Given a tweet, you are required to generate different queries from the tweet for the Google search engine to get the most relevant web content to fact-check the tweet. 
If the given tweet is not informative enough to generate a query, you should answer "none".
\end{Verbatim}
\end{footnotesize}
\caption{LLM prompt for query generation. The post information concatenated its textual content, informative descriptions of images (for posts with images), time, and the poster's name.}
\label{supp:figure:prompt_query_generation}
\end{figure}
\begin{figure}
\begin{footnotesize}
\begin{Verbatim} [breaklines,numbers=left,xleftmargin=5mm]
Given an article: 
1. Quote its paragraphs, at most two, that explicitly and completely refute the given tweet.
2. Quote its paragraphs, at most two, that implicitly refute the given tweet. 
Such paragraphs often provide the tweet's context that can imply the tweet is cherry-picking by showing the full picture. 
If the article does not have such content or is irrelevant to the tweet, you should answer `none.'
\end{Verbatim}
\end{footnotesize}
\caption{LLM prompt for evidence extraction. The article information included the article's content and published date. The post information concatenated its textual content, informative image captions (for posts with images), time, the poster's name, the poster's screen name, and the poster's description. }
\label{supp:figure:prompt_evidence_extraction}
\end{figure}
\begin{figure}
\begin{footnotesize}
\begin{Verbatim} [breaklines,numbers=left,xleftmargin=5mm]
You are required to respond to a tweet, given some facts as references. Your response should satisfy all the following requirements: 
- Your response should explain where and why the tweet is or is not misinformed or potentially
misleading. 
- You should prioritize the facts very close to the date the user tweeted, very recently, and listed at the beginning of the facts. 
- You should show the URLs that support your explanation. You should not number the URLs. 
- Your response should be informative and short. 
- Your response should start with `This tweet is.'
\end{Verbatim}
\end{footnotesize}
\caption{LLM prompt for response generation. The facts listed every piece of extracted evidence with its source link and published date. Evidence was ranked by publishers' priority and, within each priority, by relevance to the post (both from high to low), which shows to increase GPT-4's accuracy\cite{vu2023freshllms}.}
\label{supp:figure:prompt_response_generation}
\end{figure}

\begin{figure}
\begin{spacing}{0.90}
\begin{scriptsize}
\begin{Verbatim} [breaklines,numbers=left,xleftmargin=5mm]
Describe an image in an informative way. Your description should be only based on the given {short caption}, {name of each person}, and {raw text}. If the image is from social media, you should start with "A screenshot of". If the image is a quote from someone, you should start with "A quote from" followed by this person's name if there is any, then by the quoted text. If the image is an article, you should start with "An article". If the image is a photo, you should start with "A photo of". If the image is a map, you should start with "A map of". {raw text} may contain nonsense data that are unnecessarily included in the image description; however, {name of each person} is not, and if the concept in {raw text} has a conflict with that in {short caption} (e.g., "Robbie Lemos" versus "robbie leems" shown later), {raw text} is often the right one.
short caption: {a woman with glasses and a quote that says, in real life, i assure you there is no such thing as algebra}
name of each person: {Fran Lebowitz}
raw text: {"In real life, I assure you, there is no such thing as algebra."}
image description: {A quote from Fran Lebowitz, "In real life, I assure you, there is no such thing as algebra."}
short caption: {two men in suits}
name of each person: {Jim Caviezel, Michael Emerson}
raw text: {}
image description: {A photo of Jim Caviezel and Michael Emerson in suits}
short caption: {robbie leems on twitter}
name of each person: {}
raw text: {Robbie Lemos @RobbieLemos 1d I'd like to congratulate my dear friend Deep Mind on a wonderful 1st day at work today at Google. Just in time for #EarthDay2023, cheers brother! 1 2 3,790}
image description: {A screenshot of a post of Robbie Lemos, "I'd like to congratulate my dear friend Deep Mind on a wonderful 1st day at work today at Google. Just in time for #EarthDay2023, cheers brother!" The post was posted on Twitter.}
short caption: {a moose}
name of each person: {}
raw text: {Yahoo Finance @YahooFinance Typically, the stock market bottoms four to five months before a recession ends, but RBC's research details that it has bottomed as early as nine months before the end of a recession. finance.yahoo.com Could the stock market power through a recession? 'This would be rare.' 09:57 22/4/2023 3.4,011 Views 1 Retweet 1 Quote 5 Likes 1 Bookmark}
image description: {A screenshot of a post from Yahoo Finance, "Typically, the stock market bottoms four to five months before a recession ends, but RBC's research details that it has bottomed as early as nine months before the end of a recession." The post shared an article from finance.yahoo.com claiming, "Could the stock market power through a recession? 'This would be rare.'" with a picture of a moose. The post was posted at 09:57 22/4/2023.}
short caption: {a person pouring tea into a cup}
name of each person: {}
raw text: {New research reveals how coffee and tea can affect risk of early death for adults with diabetes By Sandee LaMotte, CNN Updated 7:01 PM EDT, Wed April 19, 2023 f The health benefits of tea 01:10 - Source: CNN}
image description: {An article claiming, "New research reveals how coffee and tea can affect risk of early death for adults with diabetes." It attached a picture of a person pouring tea into a cup. It was written by Sandee LaMotte, published by CNN, and updated at 7:01 PM EDT, Wed April 19, 2023.}
short caption: {two people standing next to each other with the words love is blind}
name of each person: {Nick Lachey}
raw text: {\"Love Is Blind\" co-host faceplants with a regressive line of questioning Hayley Miller MSNBC DAILY MSNBC}
image description: {An article claiming, "'Love Is Blind' co-host faceplants with a regressive line of questioning." It attached a picture of Nick Lachey and another person standing next to each other. It was written by Hayley Miller and published by MSNBC.}
short caption: {a bar graph that shows how engaged are the most followed journalists on twitter}
name of each person: {Rahul Kanwal}
raw text: {How engaged are the most-followed journalists on Twitter? Percentage of tweets from each journalist that are at-replies BDUTT 64% RealMikeWilbon 38% 34% ErinAndrews 31% stephenasmith 24% andersoncooper rahulkanwal 13% 13% ninagarcia 5% sardesairajdeep 1% BillSimmons maddow 1% 0% 10% 20% 30% 40% 50% 60% chart by mattmaldre.com}
image description: {A bar graph showing how engaged the most followed journalists, including Rahul Kanwal, are on Twitter through the percentage of tweets from each journalist that are at-replies. The chart was made by mattmaldre.com.}
short caption: {a graph showing the global defense budget by region}
name of each person: {}
raw text: {Global Defense Budgets by Region ($ Billions) $1,000 800 600 400 200 0 2020 2021 2022 2023 2024 2025 Asia-Pacific Latin America North America Sub-Saharan Africa Europe Middle East & North Africa Russia & Commonwealth of Independent States Source: Aviation Week}
image description: {A graph showing the global defense budget by region. It is from Aviation Week.}
short caption: {[IMAGE_CAPTION]}
name of each person: {[CELEBRITIES]}
raw text: {[OCR]}
image description:
\end{Verbatim}
\end{scriptsize}
\caption{LLM prompt for informative image captioning. As examples, we selected eight diverse images and manually generated their informative descriptions.}
\label{supp:figure:prompt_image2text}
\end{spacing}
\end{figure}


% Results

% Discussion

\end{document}